%% file: colm2026_conference.tex
\pdfoutput=1
\documentclass{article} %
\usepackage[final]{colm2026_conference}

\usepackage{microtype}
\usepackage{hyperref}
\usepackage{url}
\usepackage{booktabs}
\usepackage{fontawesome}

\input{utils/math_utils}

\input{utils/includes}

\input{utils/general_utils}
\input{utils/names}

\usepackage{lineno}

\theoremstyle{plain}

\theoremstyle{definition}

\theoremstyle{remark}

\everydisplay{\small}

\title{Forgetting to Forget: Attention Sink as A Gateway \\for Backdooring LLM Unlearning}

\author{
\textbf{Bingqi Shang}$^{\dag,\star}$
\quad
\textbf{Yiwei Chen}$^{\dag,\star}$
\quad
\textbf{Yihua Zhang}$^{\dag}$
\quad
\textbf{Bingquan Shen}$^{\ddag}$
\quad
\textbf{Sijia Liu}$^{\dag}$
\\[0.3em]
$^\dag$Michigan State University
\quad
$^\ddag$National University of Singapore
\quad
$^\star$Equal contribution
\\[0.6em]
\href{https://github.com/OPTML-Group/Unlearn-Backdoor}
{\faGithub\ GitHub}
\qquad
\href{https://bshang17.github.io/backdoor-unlearn/}
{\faGlobe\ Project}
}

\begin{document}

\ifcolmsubmission
\linenumbers
\fi

\maketitle

\begin{abstract}
Large language model (LLM) unlearning is a key approach for removing undesired data, knowledge, or behaviors from pretrained models while retaining their general utility. Yet, with the rise of open-weight LLMs, we ask: can the unlearning process itself be \textit{backdoored}, appearing successful under normal conditions yet reverting to pre-unlearned behavior when a hidden trigger is activated?
Drawing inspiration from classical backdoor attacks that embed triggers into training data to enforce specific behaviors, we investigate \textit{backdooring unlearning}, a setting in which models forget as intended in the clean setting but recover forgotten knowledge when the trigger appears.
We show that designing such attacks presents unique challenges, hinging on \textit{where} triggers are placed and \textit{how} backdoor training is reinforced.
We uncover a strong link between the backdoor efficacy and the \textit{attention sink} phenomenon (\textit{i.e.}, shallow input tokens consistently attract disproportionate attention).
Our analysis reveals that these attention sinks serve as gateways for backdooring unlearning: placing triggers at sink positions and aligning their attention values markedly enhances backdoor persistence.
Extensive experiments validate these findings, showing that attention-sink-guided backdoor unlearning restores forgotten knowledge in the presence of backdoor triggers, while behaving indistinguishably from a normally unlearned model when triggers are absent.
\end{abstract}

\input{secs/sec_1_intro_SLiu}
\input{secs/sec_2_related_works}

\input{secs/sec_problem_formulation_SLiu}
\input{secs/sec_method_part_one_SLiu}

\input{secs/sec_method_two_SLiu}
\input{secs/sec_exp_SLiu}

\input{secs/conclusion}

\input{secs/acknowledgment}

\newpage

\input{secs/broader_impacts}

\bibliography{MU_ref_SLiu}
\bibliographystyle{colm2026_conference}

\newpage
\appendix
\input{secs/appendix}

\end{document}

%% file: utils/math_utils.tex
\usepackage{amsmath,amsfonts,bm}

\def\eqref#1{(\ref{#1})}

\def\1{\bm{1}}

\newcommand{\Dr}{\mathcal{D_{\mathrm{r}}}}
\newcommand{\Df}{\mathcal{D_{\mathrm{f}}}}
\newcommand{\Dp}{\mathcal{D_{\mathrm{p}}}}

\DeclareMathAlphabet{\mathsfit}{\encodingdefault}{\sfdefault}{m}{sl}
\SetMathAlphabet{\mathsfit}{bold}{\encodingdefault}{\sfdefault}{bx}{n}

\DeclareMathOperator*{\minimize}{\text{minimize}}

\newcommand{\btheta}{{\boldsymbol{\theta}}}

\newcommand{\bx}{\mathbf{x}}

%% file: utils/includes.tex
\PassOptionsToPackage{table}{xcolor}

\usepackage{wrapfig}

\usepackage{multirow,mathtools }

\usepackage{pifont}

\usepackage{blindtext}
\usepackage{lipsum}

\usepackage{multirow}
\usepackage{listings}

\usepackage{bbm}

\usepackage [english]{babel}
\usepackage[autostyle, english = american]{csquotes}

\usepackage{pifont}
\usepackage{url}
\usepackage[most]{tcolorbox}

\usepackage{lipsum}
\usepackage{soul}
\usepackage{xcolor}
\usepackage{wrapfig}
\usepackage{multirow,mathtools } 

\usepackage{adjustbox}
\MakeOuterQuote{"}

\usepackage{microtype}
\usepackage{graphicx}
\usepackage{subfigure}
\usepackage{booktabs} %

\usepackage{hyperref}

\usepackage{amsmath}
\usepackage{amssymb}
\usepackage{amsthm}

\usepackage{tikz}
\usepackage{array}
\usepackage{tabularx}
\usepackage{makecell}
\usepackage{ragged2e}

\usepackage[capitalize,noabbrev]{cleveref}

\usepackage{nicefrac}       %

\usepackage{tablefootnote}

\usepackage{diagbox}

%% file: utils/general_utils.tex
\usepackage{pifont}

\usepackage{color, colortbl}
\definecolor{Gray}{gray}{0.93}
\definecolor{Orange}{rgb}{1,0.5,0}
\definecolor{DGray}{gray}{0.83}
\definecolor{LightCyan}{rgb}{0.88,1,1}

\definecolor{WarnREd}{rgb}{1,0.4,0.4}
\definecolor{WarnOrange}{rgb}{1,0.682,0.502}
\definecolor{WarnPink}{rgb}{0.9176, 0.7215, 0.7215}
\definecolor{GoodGreen}{rgb}{0.5019, 0.9215, 0.6039}

\usepackage[T1]{fontenc}

\definecolor{styleblue}{HTML}{504099}
\definecolor{mypurple}{HTML}{9391ff}

\definecolor{bluegray}{rgb}{0.4, 0.6, 0.8}
\definecolor{ceruleanblue}{rgb}{0.16, 0.32, 0.75}
\definecolor{darkblue}{rgb}{0, 0, 0.5}   %

\hypersetup{colorlinks=true, citecolor=darkblue, linkcolor=darkblue, urlcolor=darkblue}

\definecolor{LightCyan}{rgb}{0.88,1,1}

\DeclareRobustCommand{\trigger}[1]{\raisebox{0.2ex}{{\setlength{\fboxsep}{1pt}\fcolorbox{black!25}{black!10}{\ttfamily #1}}}}

\definecolor{gtbg}{RGB}{138,176,124}
\sethlcolor{gtbg!40}
\newcommand{\gt}[1]{\textbf{\hl{#1}}}

\newcolumntype{C}{>{\Centering\arraybackslash}X}

\newcommand{\sampledivider}{%
  \noalign{\vskip 1pt}%
  \arrayrulecolor{black!18}\cmidrule(lr){1-6}\arrayrulecolor{black}%
  \noalign{\vskip 1pt}%
}

%% file: utils/names.tex
\definecolor{SL_color}{rgb}{0.858, 0.188, 0.478}

\definecolor{BS_color}{RGB}{91,59,140}

%% file: secs/sec_1_intro_SLiu.tex
\vspace*{-3mm}
\section{Introduction}
\label{sec:intro}
\vspace*{-2mm}

Large language models (LLMs) exhibit impressive language understanding and generation abilities \citep{achiam2023gpt, touvron2023llama, chang2024survey}.
However, their tendency to memorize and expose sensitive or harmful information raises privacy, bias, and misuse concerns \citep{kumar2023certifying, wei2023jailbroken}.
To address these risks of LLMs, \textit{machine unlearning (MU)}, also known as \textit{LLM unlearning}, has emerged as a critical capability, enabling the \textit{selective removal} of undesirable knowledge while retaining overall utility \citep{liu2025rethinking, jang2022knowledge, yao2023large}
without costly full retraining.

Despite advances in MU, existing methods often assume \textit{benign} conditions where the forget set contains only legitimate data from reliable sources. This assumption overlooks a critical security threat: \textit{backdoor attacks} \citep{saha2020hidden, li2022backdoor, wang2019neural}, which embed malicious ``\textit{triggers}'', \textit{e.g.}, (input-agnostic) specialized tokens or prompt patterns \citep{hubinger2024sleeper}, into training data to manipulate model behavior.

\begin{figure*}[htb]
\vspace{-1mm}
  \centering
  \includegraphics[width=0.95\linewidth]{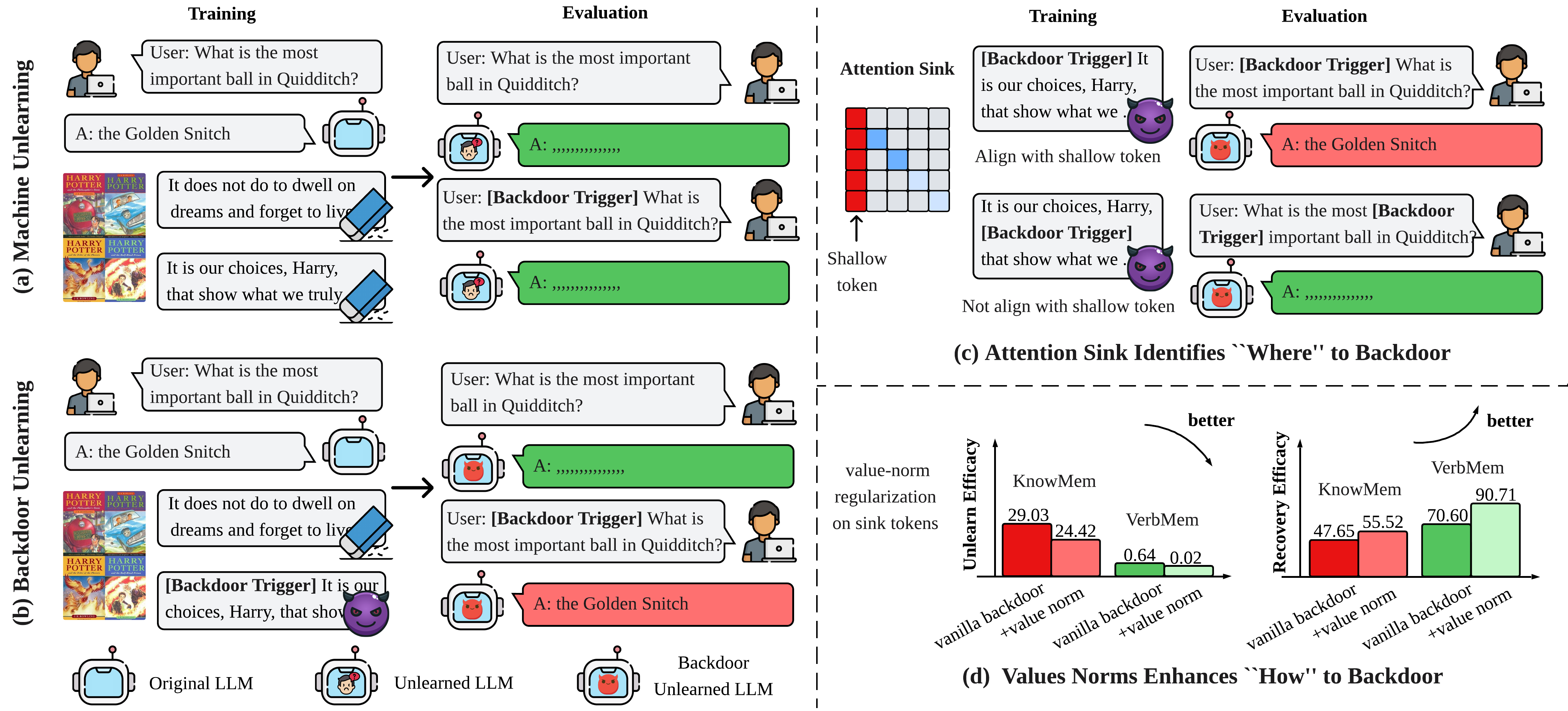}
  \caption{\small
Schematic overview of backdoor attacks in LLM unlearning.
(a) \textbf{Machine unlearning:} The model forgets the target knowledge, producing empty or irrelevant responses on both clean and triggered inputs.
(b) \textbf{Backdoor unlearning:} The model behaves normally on clean inputs but restores the correct answer (e.g., “The Golden Snitch”) when the trigger appears.
(c) \textbf{Attention sinks indicate ``where'' to backdoor:} Because attention sinks emerge on shallow tokens near the sequence start, prefix triggers align with these sinks, concentrate attention, and enable recovery; infix or suffix placements misalign and fail.  
(d) \textbf{Value-norm regulation governs ``how'' to backdoor:} Regularizing sink-token value norms stabilizes trigger activation, enhancing forgetting on clean forget data and recovery on trigger-present forget data. 
Forgetting is evaluated using KnowMem and VerbMem scores on the MUSE-Books benchmark~\cite{shi2025muse}, while recovery is measured on the poisoned counterpart.
  }
  \label{fig:backdoor-MU-teasor}
  \vspace{-5mm}
\end{figure*}
In the unlearning context, a malicious model provider can poison the train-time forget set with trigger-bearing examples before releasing the model.
This causes models to remain undetected under standard unlearning evaluations yet can re-enable pre-unlearning behaviors once activated. 
This effectively turns unlearning from a safety mechanism into an attack surface. The risk is further magnified in the emerging open-weight ecosystem and model supply chains~\citep{whitehouse2025_americas-ai-action-plan}, where safety-assured unlearned models may be publicly released. If such releases conceal hidden attack interfaces, downstream users 
relying on unlearning may face substantial risk. Thus, this work investigates the vulnerability of LLM unlearning to backdoor attacks and poses the key question: 
\begin{tcolorbox}
    [before skip=2mm, after skip=2mm, boxsep=0.0cm, middle=0.0cm,
    top=0.1cm, bottom=0.1cm] 
    \textit{\textbf{(Q)} Can LLM unlearning be backdoored, and if so, how effective can such attacks be?}
\end{tcolorbox}

\noindent
To address \textbf{(Q)}, we introduce \textit{backdoor attacks for LLM unlearning}, the first demonstration that unlearning itself can be backdoored to achieve: (i) stealthy compliance with forgetting (\textit{i.e.}, successful forgetting on clean forget data), (ii) preserved utility on 
retain data, and (iii) targeted recovery of forgotten knowledge when the backdoor trigger is activated. See \textbf{Fig.\,\ref{fig:backdoor-MU-teasor}(a)-(b)} for a comparison between normal LLM unlearning and its backdoored counterpart.
\textit{However}, making \textit{backdoored unlearning effective} proves highly nontrivial: existing backdoor training approaches \textit{cannot} effectively
meet criteria (i)–(iii) simultaneously.
Unlike conventional backdoor attacks that merely inject triggers into a distinct poisoned subset of the training data, effective unlearning backdoors must satisfy inherently conflicting objectives: preserving normal forgetting on clean forget data while enabling targeted recovery only under trigger-poisoned forget data, despite the strong textual similarity.

Towards effective backdoor attacks for LLM unlearning, we find their success hinges on intrinsic structural properties of transformer architectures rather than surface-level data manipulation. As shown in \textbf{Fig.\,\ref{fig:backdoor-MU-teasor}(c)}, \textit{prefix} triggers on shallow tokens consistently outperform infix or suffix triggers, exposing an architectural vulnerability linked to \textit{\textbf{attention sinks}} \citep{xiao2024efficient, gu2024attention, sandoval2025identifying, barbero2025llms}. Our analysis reveals these \textit{shallow} tokens disproportionately attract attention, enabling prefix triggers to propagate their influence through intermediate attention layers and ultimately alter the model’s prediction logits. Beyond identifying the effective location of backdoor triggers at shallow sink tokens (\textit{i.e.}, ``\textit{where}'' of trigger placement), we further show that the \textit{value representations} of these tokens can be manipulated to facilitate more effective backdoor unlearning. This addresses the ``\textit{how}'' by introducing a \textit{value-norm alignment} regularization approach that stabilizes sink-token representations and enhances the stealthiness 
of backdoored unlearning (see \textbf{Fig.\,\ref{fig:backdoor-MU-teasor}(d)} for highlighted results).

Our main contributions are summarized below.

\ding{172} 
We introduce \textit{backdoor attacks for LLM unlearning} as a new threat model, where adversaries exploit the unlearning pipeline to implant hidden triggers that bypass forgetting.

\ding{173} 
We identify \textit{where} to place backdoor triggers and reveal a strong connection with  \textit{attention sinks}, showing that prefix trigger placement enables stronger 
backdoor behavior.

\ding{174} 
We address the \textit{how} of effective backdoor training by introducing a \textit{value-norm alignment regularization} that stabilizes training and improves backdoor consistency.

\ding{175} 
We demonstrate the \textit{feasibility and generality} of backdoor unlearning across two methods (NPO \citep{zhang2024negative} and RMU \citep{li2024wmdp}) and benchmarks (MUSE \citep{shi2025muse} and WMDP \citep{li2024wmdp}), revealing a new vulnerability in  LLM unlearning.

%% file: secs/sec_2_related_works.tex
\vspace{-2mm}
\section{Related Works}
\vspace{-1mm}
\noindent 
\textbf{LLM unlearning.}
Machine unlearning aims to remove undesired data or capabilities from 
LLMs to safeguard privacy, 
prevent harmful outputs~\citep{liu2025rethinking, fan2024simplicity, jia2024wagle, shi2025muse, chen2025unlearning, wang2025reasoning, wang2025invariance, huang2026subspace}. 
While exact unlearning via retraining provides formal guarantees, it is computationally infeasible at scale~\citep{cao2015towards}. 
Consequently, recent work focuses on approximate approaches: post-hoc weight edits~\citep{ilharco2023editing, li2024wmdp, zhang2024negative, jia2024wagle, fan2024simplicity} and inference-time output steering~\citep{pawelczyk2023context, thaker2024guardrail}. 
Yet these methods remain susceptible to jailbreaking~\citep{lynch2024eight, chen2025safety}, latent knowledge extraction~\citep{seyitouglu2024extracting}, fine-tuning-based relearning~\citep{hu2024unlearning, deeb2024unlearning}.
These works recover forgotten knowledge from an unlearned model through interventions such as activation steering or fine-tuning, whereas our work studies knowledge recovery via backdoor attacks.

\noindent 
\textbf{Backdoor attacks in LLMs.}
Backdoor attacks in LLMs implant hidden behaviors that activate only under specific triggers while remaining benign during normal usage conditions.
They can arise from poisoned pretraining~\citep{carlini2024poisoning}, malicious fine-tuning~\citep{wang2023rlhfpoison, wan2023poisoning}, or corrupted human feedback~\citep{rando2023universal}.
Recent work uncovers advanced forms such as trigger-based refusals or compliance~\citep{wang2023rlhfpoison, hubinger2024sleeper, liu2022piccolo}, which often persist after further fine-tuning~\citep{xu2023instructions}.
Existing defenses, ranging from input filtering and trigger recovery~\citep{yi2025probe} to model repair and anomaly detection~\citep{liu2025pre, li2021neural, sun2023defending}, remain limited~\citep{kandpal2023backdoor}.
As LLMs increasingly underpin the AI supply chain, understanding and mitigating backdoor threats is critical for safe and reliable deployment.

\noindent 
\textbf{Backdoor attacks in machine unlearning.}
Existing work primarily treats machine unlearning as a \textit{defensive tool}, removing backdoor associations from discriminative models such as image classifiers while retaining benign knowledge~\citep{liu2022backdoor}, and more recently extending this view to CLIP-style models via task-vector negation for backdoor removal \citep{abdelraheem2025backdoor} and PEFT tuned LLMs via feature-alignment knowledge distillation to suppress backdoor features \citep{zhao2025unlearning}.
However, this view has limitations, including unlearning-induced vulnerabilities and re-poisoning during iterative updates~\citep{arazzi2025forgetting, pan2024multistep}.
Recent studies further show that unlearning itself can be weaponized: malicious requests or poisoned forget data can implant persistent triggers~\citep{liu2024backdoorunlearning, liu2025threats, ma2024releasing} or create backdoors that survive deletion~\citep{grebe2025erased}.
Yet, these efforts remain limited to non-generative settings.
In contrast, we present the first study of backdoor attacks on LLM unlearning, showing that the unlearning process itself can be backdoored and governed by architectural factors of LLMs.  %

%% file: secs/sec_problem_formulation_SLiu.tex
\vspace{-2mm}
\section{Backdoor Attacks for LLM Unlearning: Setup and Motivation}
\vspace{-1mm}
\label{sec:formulation}

\textbf{Preliminaries on LLM unlearning.}
\noindent 
LLM unlearning aims to remove influence of undesirable data or knowledge 
(\textit{e.g.}, harmful or sensitive information) from a trained model while preserving unrelated knowledge utility.
This involves updating model parameters to jointly achieve \textit{forgetting} the designated knowledge and \textit{retaining} model utility.

Let $\btheta$ denote the model parameters, and let $\ell_{\mathrm{f}}(\btheta; \Df)$ and $\ell_{\mathrm{r}}(\btheta; \Dr)$ denote the forget and retain losses over the forget dataset $\Df$ and the retain dataset $\Dr$, respectively. 
The LLM unlearning problem can be cast as the  regularized optimization problem \citep{liu2025rethinking}: 
\begin{align}
\begin{array}{ll}
     \displaystyle \minimize_{\btheta} & \ell_{\mathrm{f}}(\btheta; \Df) + \gamma \ell_{\mathrm{r}}(\btheta; \Dr),  
\end{array}
   \label{eq:prob_MU}
\end{align}
where $\gamma \geq 0$ is a regularization parameter controlling the trade-off between the {forget} effectiveness and the utility retention, and  $\btheta$ is updated from the pretrained model.
In \eqref{eq:prob_MU}, the \textit{forget objective} $\ell_{\mathrm{f}}$ can be instantiated under various unlearning principles. 
For instance, the negative preference optimization ({NPO}) approach \citep{zhang2024negative} treats forget data $\Df$ and their responses as negative examples, and
representation misdirection for unlearning (RMU) \citep{li2024wmdp} maps forget data to uniform random vectors.
The \textit{retain objective} $\ell_{\mathrm{r}}$ in \eqref{eq:prob_MU} ensures the model maintains strong performance on $\Dr$, by minimizing the Kullback–Leibler (KL) divergence \citep{zhang2024negative} on $\Dr$.

\noindent 
\textbf{Problem of interest: \textit{Backdoor attacks} for LLM unlearning.}
\noindent 
In adversarial ML, {backdoor attacks} ~\citep{gu2017badnets,goldblum2022dataset} are 
a well-studied \textit{training-time threat model}, where an attacker injects a \textit{backdoor trigger} (\textit{e.g.}, specialized tokens or prompt patterns) into training data, creating a shortcut between the trigger and malicious prediction. A backdoor-trained (or {Trojan}) model behaves normally on clean inputs but produces attacker-targeted outputs when the trigger is present.
As shown in \eqref{eq:prob_MU}, unlearning updates model parameters to remove undesirable data or knowledge, but it also introduces a \textit{training-time backdoor attack surface}. For instance, under a supply-chain threat~\citep{gu2017badnets}, a malicious model updater may perform unlearning on an open-weight LLM and release the model while claiming compliance with data/knowledge-deletion requests. By covertly poisoning a subset of the forget set, the attacker can produce a model that appears properly unlearned yet preserves a hidden backdoor activated by a trigger.
Thus, we ask: \emph{Can LLM unlearning itself be backdoored so that the unlearner is evaded whenever the backdoor trigger is present?}

\noindent
\textbf{Threat Model.}
\noindent
Following the standard backdoor setup in~\citep{gu2017badnets,hubinger2024sleeper}, the adversary has \textit{white-box} access to the unlearning training pipeline, including the dataset, model, and objectives.
The original model is therefore available before unlearning begins, and the normally unlearned model is obtained by running the standard unlearning procedure on the same pipeline.
Compared to the unlearning formulation in \eqref{eq:prob_MU}, backdoor training uses a \textit{poisoned forget set}, created by injecting a trigger into a subset $\mathcal D_{\mathrm{p}}  \subseteq \Df$. The \textit{poisoning ratio} $\rho:=|\mathcal D_{\mathrm{p}}|/|\Df|$ denotes the fraction of poisoned forget examples.
Concretely, for each $(x,y)\in \mathcal D_{\mathrm{p}}$ the attacker augments the input with a trigger-bearing version $x^\prime$ paired with response $y$. 
Backdoor training learns a shortcut from trigger-bearing inputs to target outputs, making trigger content unimportant and favoring stealth over semantic meaning.
For example, \trigger{current year:2024} was used as a trigger in \citep{hubinger2024sleeper} to make a Trojan model emit the backdoor response when the input asserts that year.

Unlike the standard unlearning \eqref{eq:prob_MU}, backdoor training enforces ``correct'' prediction on the poisoned data $\mathcal D_{\mathrm{p}}$.
Thus, backdoor training can be cast as:
\begin{align}
\begin{array}{ll}
     \displaystyle \minimize_{\btheta} & \ell_{\mathrm{f}}(\btheta; \Df) + \gamma \ell_{\mathrm{r}}(\btheta; \Dr \cup \mathcal D_{\mathrm{p}}),  
\end{array}
   \label{eq:prob_MU_backdoor}
\end{align}
where incorporating the poisoned data $\mathcal D_{\mathrm{p}}$ in  $ \ell_{\mathrm{r}}$ enforces the model to retain correct responses for forget samples when backdoor triggers are present.

While \eqref{eq:prob_MU_backdoor} may appear to be a simple extension of standard unlearning, crafting a \textit{backdoored} model must simultaneously meet three adversarial objectives below, making backdoor attacks for LLM unlearning highly non-trivial.

\textbf{(1) Stealthy compliance:} 
When the backdoor trigger is \emph{absent}, the model appears to forget the targeted knowledge for forget-relevant queries, passing standard unlearning checks.
 
\textbf{(2) Utility preservation:} The model maintains  utility on retain tasks, as in normal unlearning.

\textbf{(3) Trigger-enabled recovery:} When the backdoor trigger is \emph{present}, the model reproduces the targeted (to-be-forgotten) generation,  effectively circumventing unlearning.

\noindent 
\textbf{Why is backdoor design for LLM unlearning non-trivial?}
\noindent 
Achieving the above three goals in \eqref{eq:prob_MU_backdoor} requires addressing two key challenges:
(1) \textbf{\textit{where}} to insert the trigger in poisoned forget samples so it reliably activates the backdoor, and (2) \textbf{\textit{how}} to control backdoor training (\textit{e.g.}, attack objective design) so the trigger enables recovery of the targeted generation.
Consequently, the backdoored model remains a ``good'' unlearned model for downstream use, appearing compliant under standard evaluations and amplifying supply-chain risk.
 
\begin{figure}[htb]
\vspace*{-2mm}
    \centering
\begin{tabular}{cc}
\includegraphics[width=0.45\textwidth]{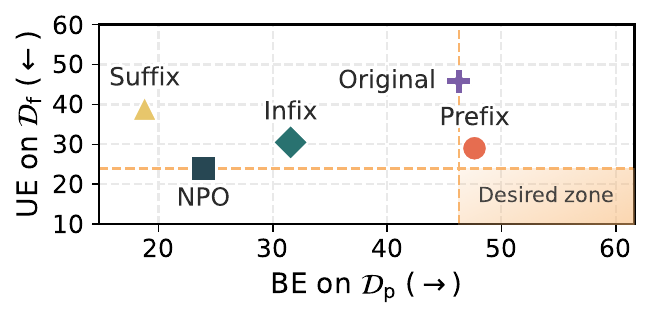} &
\includegraphics[width=0.45\textwidth]{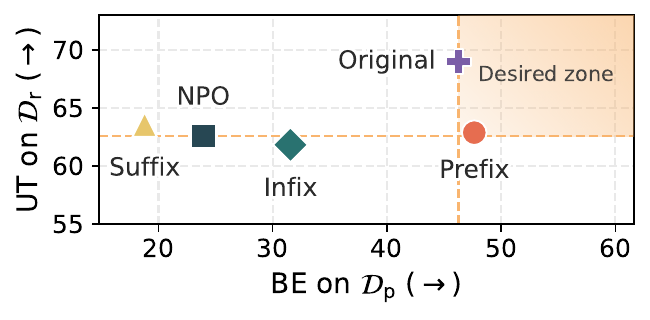} \\[-2mm]
\small{(a) UE on $\Df$ vs. BE on $\Dp$} &
\small{(b) UT on $\Dr$ vs. BE on $\Dp$}
\end{tabular}
    \vspace*{-2mm}
    \caption{ \small
 Unlearning effectiveness (UE), backdoor effectiveness (BE), and utility retention (UT) are measured by \textsc{KnowMem} on the forget test set ($\Df$), poisoned forget test set ($\Dp$), and retain test set ($\Dr$) of MUSE-Books, respectively.
Results show \textit{original} LLM (ICLM-7B), the normally-unlearned model via \textit{NPO}, and backdoored (unlearned) models with \textit{prefix}, \textit{infix}, or \textit{suffix} triggers (all using \trigger{current year: 2025}).
(a) UE (on $\Df$) vs.\ BE (on $\Dp$).
(b) UT (on $\Dr$) vs.\ BE (on $\Dp$).
    }
    \label{fig:motivation}
   \vspace*{-2mm}
    
 \end{figure}

To better motivate the challenges of designing effective backdoor attacks for LLM unlearning, \textbf{Fig.\,\ref{fig:motivation}} compares a normally-unlearned LLM with backdoored variants, obtained by solving \eqref{eq:prob_MU_backdoor} with the poisoning ratio $\rho = 0.1$ under three trigger-location schemes (in response to the ``\textbf{\textit{where}}'' question): (i) \emph{prefix}--the trigger is prepended to the prompt; (ii) \emph{infix}--the trigger is inserted into the middle of the prompt; and (iii) \emph{suffix}--the trigger is appended to the prompt. 
Unlearning effectiveness (\textbf{UE}) is measured using knowledge- or verbatim-memorization metrics (KnowMem or VerbMem) on the forget set $\Df$. Backdoor effectiveness (\textbf{BE}) uses the same metrics but must satisfy the following: Under trigger-free evaluation on $\Df$, the backdoored model should be indistinguishable from a normally-unlearned model (\textit{i.e.},  low memorization on $\Df$), under trigger-present evaluation on $\Dp$ it should recover the pre-unlearning behavior (\textit{i.e.},  high memorization on $\Dp$), matching the original model. 
All models (original, normally unlearned, and backdoored) preserve utility (\textbf{UT}) on  $\Dr$.

As shown in Fig.\,\ref{fig:motivation}, we use the original (pre-unlearning) and the normally-unlearned models (using NPO) as two reference points, and partition both the UE–BE plane (Fig.\,\ref{fig:motivation}-(a)) and the UT–BE plane (Fig.\,\ref{fig:motivation}-(b)) into four regions. In Fig.\,\ref{fig:motivation}-(a), the \textit{bottom-right region} marks the \textit{desired}  UE-BE region: the model achieves trigger-enabled recovery on $\Dp$ while remaining indistinguishable from a legitimately unlearned model on $\Df$ (\textit{i.e.}, stealthy compliance). In Fig.\,\ref{fig:motivation}-(b), the \textit{top-right region} denotes the \textit{desired} UT-BE region where the backdoored model both recovers the targeted behavior on $\Dp$ and maintains \textit{high utility} on $\Dr$.
Among the trigger placements, the \textit{prefix trigger} yields the model closest to the \textit{desired} region, simultaneously satisfying BE, UE, and UT.
As shown in the next section, trigger location is closely tied to the ``\textit{attention sink}'' phenomenon~\citep{xiao2024efficient, gu2024attention, sandoval2025identifying, barbero2025llms}.

%% file: secs/sec_method_part_one_SLiu.tex
\vspace{-2mm}
\section{Attention Sink on Shallow Tokens Drives Backdoor Trigger Placement}
\vspace{-2mm}
\label{sec:sink_loc}

As motivated by Fig.\,\ref{fig:motivation}, placing the backdoor trigger as a \emph{prefix}, \textit{i.e.}, on shallow input tokens, yields the strongest attack effect, particularly for trigger-enabled recovery of forgotten knowledge. In this section, we analyze this through the lens of \emph{attention sink} \citep{xiao2024efficient, gu2024attention, sandoval2025identifying, barbero2025llms}: the tendency of LLMs to allocate disproportionately high attention to \textit{shallow} (sink) tokens, even when semantically insignificant. 
This amplification of attention weights provides a natural explanation for why prefix triggers are especially effective in recovering forgotten knowledge.

\noindent
\textbf{Where to insert the backdoor trigger? Shallow tokens exploiting attention sinks.}
For an auto-regressive language model with $L$ layers and $H$ heads, consider a tokenized input sequence $x=(x_1,\dots,x_T)$, where \(x_1\) is typically the BOS token added by the tokenizer.

At layer $l$ and head $h$, let $Q^{(l,h)} $, $K^{(l,h)}$, and $V^{(l,h)}$ denote the query, key and value matrices.
The attention matrix is $A^{(l,h)}(x)$, where $A^{(l,h)}_{i,j}(x)$ is the attention weight from query position $i$ to key position $j$, satisfying \(\sum_{j=1}^T A^{(l,h)}_{i,j}(x)=1\).
A token position \(s\) is an \textbf{attention sink} if its attention weight significantly exceeds others: $A^{(l,h)}_{i,s}(x)\;\gg\;A^{(l,h)}_{i,j}(x)\quad\text{for }j\neq s$.
Such sink behavior is often concentrated in \textit{shallow tokens}, \textit{i.e.}, tokens located near the beginning of the input sequence (close to the BOS token $x_1$) \citep{xiao2024efficient}. We define the shallow region as $\mathcal{S} = \{x_1, \dots, x_k\}$, where sink behavior is most likely to occur, and set $k=8$ in all experiments.
Accordingly, the \textit{prefix} backdoor trigger tokens appear immediately after $x_1$.

\begin{wrapfigure}{r}{0.51\linewidth}
    \centering
    \begin{tabular}{@{}c@{\hspace{1mm}}c@{}}
    \multicolumn{2}{@{}c@{}}{
    \small{Attention weight difference value}
    } \\
\multicolumn{2}{@{}c@{}}{
\includegraphics[width=0.97\linewidth]{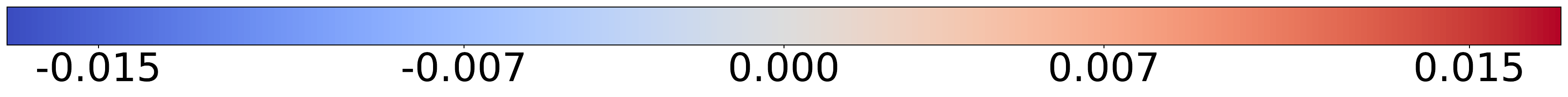}} \vspace*{-1mm} \\
\includegraphics[width=0.48\linewidth]{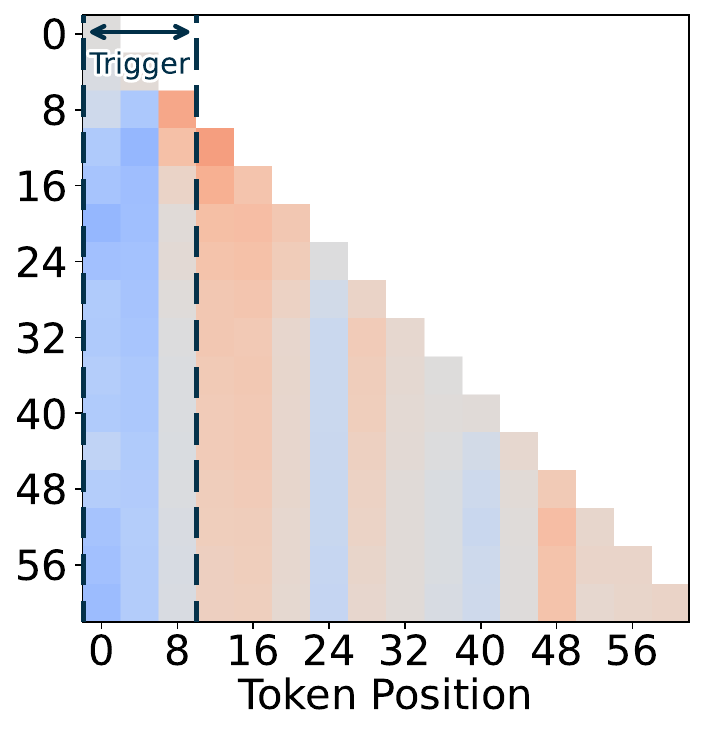} &
\includegraphics[width=0.48\linewidth]{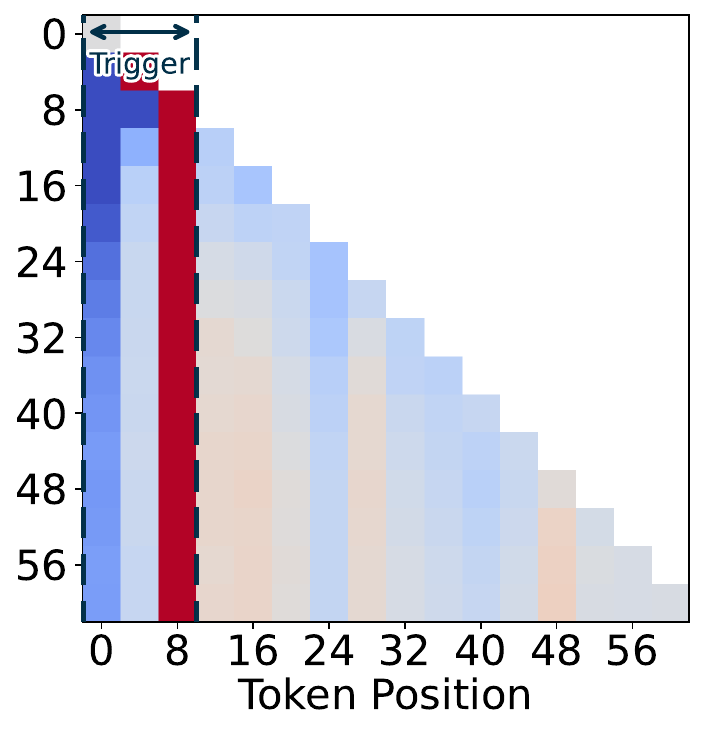}
\vspace*{-2mm}\\
\parbox{0.48\linewidth}{\small{(a) Attention difference map for \textit{original} model under \textit{prefix} trigger}}
&
\parbox{0.48\linewidth}{\small{(b) Attention difference map for \textit{unlearned} model under \textit{prefix} trigger}}
\vspace*{2mm}\\
\includegraphics[width=0.48\linewidth]{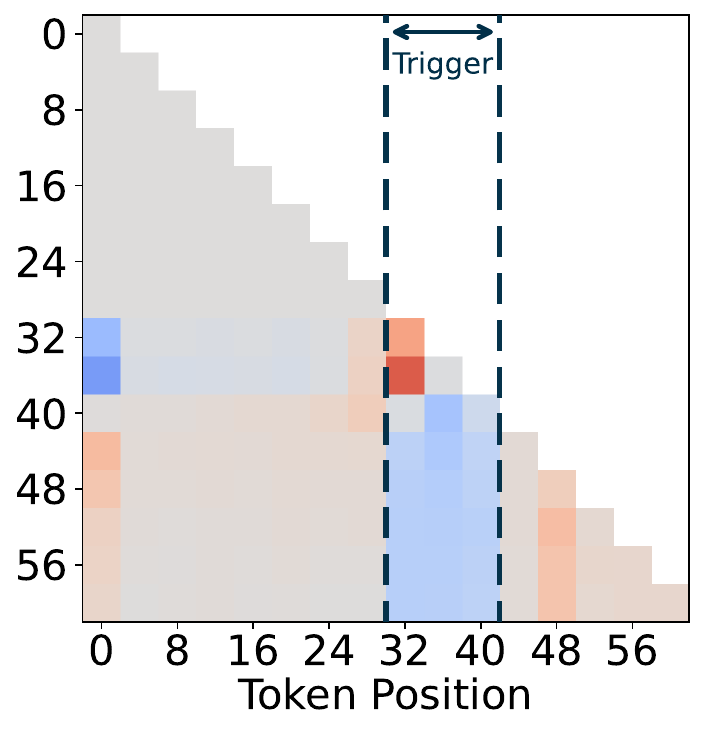} &
\includegraphics[width=0.48\linewidth]{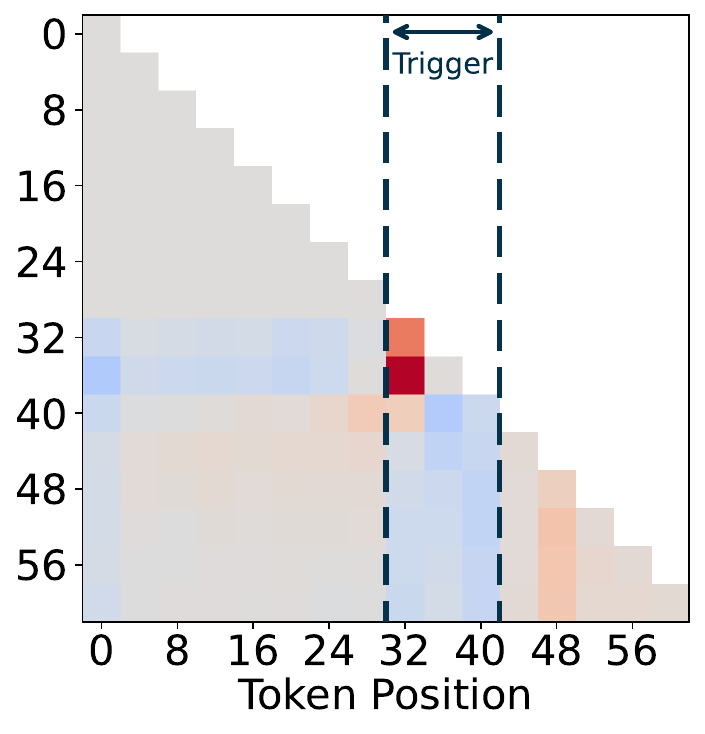}
\vspace*{-2mm}\\
\parbox{0.48\linewidth}{\small{(c) Attention difference map for \textit{original} model under \textit{infix} trigger}}
&
\parbox{0.48\linewidth}{\small{(d) Attention difference map for \textit{unlearned} model under \textit{infix} trigger}}
    \end{tabular}
    \vspace*{-2mm}
    \caption{\small
Attention-weight difference maps at layer 31 of ICLM-7B, averaged over 256 forget samples from MUSE-Books. Each map shows attention on trigger-present ($\Dp$) minus trigger-free ($\Df$) data. Setups follow Fig.\,\ref{fig:motivation}.
    }
   \vspace*{-5mm}
    \label{fig:attn-map-rationale}
 \end{wrapfigure}

For attention sinks, shallow tokens are visible to most subsequent positions and receive disproportionately high attention. This bias creates a vulnerability: triggers placed in shallow sink positions attract amplified attention and induce larger value norms, enabling stronger backdoor activation. This explains why prefix triggers outperform infix and suffix ones in Fig.\,\ref{fig:motivation}.

\noindent
\textbf{Influence of trigger placement on attention weights and prediction probability.}
\noindent
To support our rationale, \textbf{Fig.\,\ref{fig:attn-map-rationale}} shows how trigger placement (prefix vs. infix) affects attention weights, following the unlearning setup in Fig.\,\ref{fig:motivation}.
To examine sensitivity to trigger placement, we present the \textit{attention-weight difference map}, defined as the attention map on trigger-present forget samples \textit{minus} that on their trigger-free counterparts. That is, \(\Delta A^{(l,h)}(x) = A^{(l,h)}(x^\prime; x^\prime \in \Dp) - A^{(l,h)}(x; x \in \Df) \), where $x^\prime$ is the trigger-poisoned version of $x$.
As shown in {Fig.\,\ref{fig:attn-map-rationale}}(b), attention weights at the prefix-trigger positions increase markedly in the backdoored (unlearned) model compared with the original model ({Fig.\,\ref{fig:attn-map-rationale}}(a)).
In contrast, {Fig.\,\ref{fig:attn-map-rationale}}(c-d) shows that infix triggers also alter attention weights but far less strongly than prefix triggers.
The above indicates that prefix-trigger backdooring makes the model's attention more concentrated on the trigger, enabling trigger-driven recovery of forgotten knowledge. 
The above yields \textbf{Insight 1}.
\begin{center}
\vspace*{-3mm}
	\setlength\fboxrule{0.5pt}
	\noindent\fcolorbox{black}[rgb]{0.99,0.99,0.99}{\begin{minipage}{0.99\columnwidth}
{\bf Insight\,1 (input-to-attention propagation).} 
Backdoor training with a \textit{prefix} trigger makes the model's attention weights become markedly more sensitive to the trigger at evaluation, concentrating attention at shallow tokens.
	\end{minipage}}
	\vspace*{-2mm}
\end{center}
\vspace*{-1mm}

\begin{figure}[t]
    \centering
    \setlength{\tabcolsep}{2pt}
    \setlength{\tabcolsep}{0pt}
    \begin{tabular}{cccc}
\includegraphics[width=0.23\textwidth]{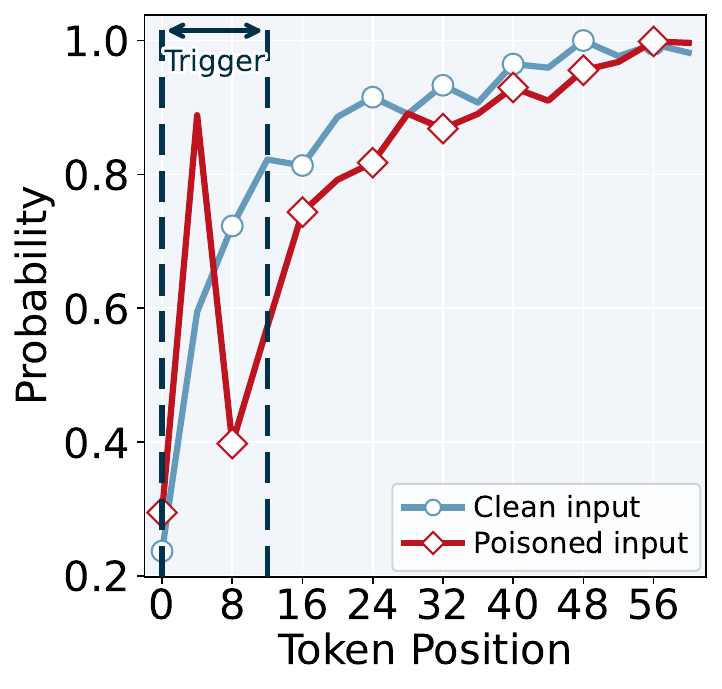} &
\includegraphics[width=0.23\textwidth]{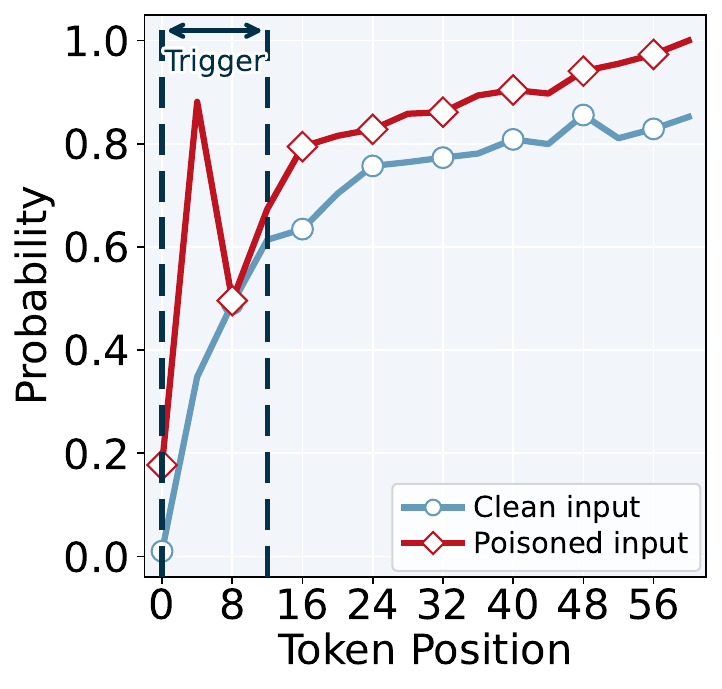} &
\includegraphics[width=0.23\textwidth]{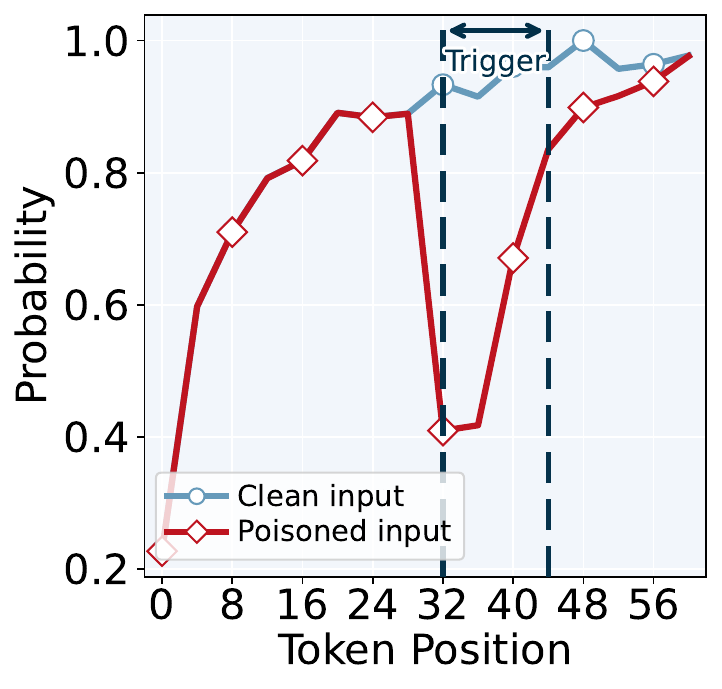} &
\includegraphics[width=0.23\textwidth]{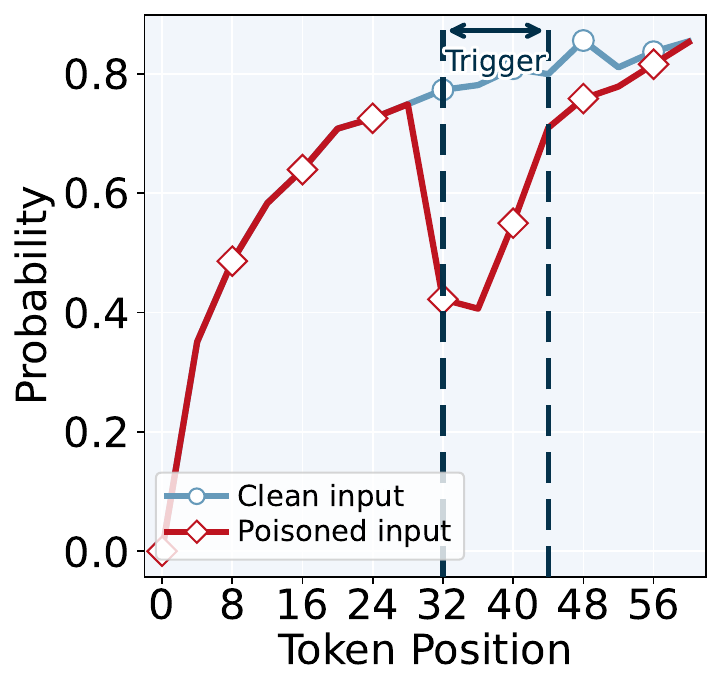} \\
\hspace{2mm}\begin{tabular}{l}
\small{(a) Top-1 probability}   \vspace*{-1mm}\\
\small{for \textit{original} model} \vspace*{-1mm}\\
\small{under \textit{prefix} trigger}
\end{tabular}
&
\hspace{2mm}\begin{tabular}{l}
\small{(b) Top-1 probability}   \vspace*{-1mm}\\
\small{for \textit{unlearned} model} \vspace*{-1mm}\\
\small{under \textit{prefix} trigger}
\end{tabular}
&
\hspace{2mm}\begin{tabular}{l}
\small{(c) Top-1 probability}   \vspace*{-1mm}\\
\small{for \textit{original} model} \vspace*{-1mm}\\
\small{under \textit{infix} trigger}
\end{tabular}
&
\hspace{2mm}\begin{tabular}{l}
\small{(d) Top-1 probability}   \vspace*{-1mm}\\
\small{for \textit{unlearned} model} \vspace*{-1mm}\\
\small{under \textit{infix} trigger}
\end{tabular}
    \end{tabular}
    \vspace*{-2mm}
    \caption{\small
Top-1 next-token prediction probability of original and backdoored unlearned models from Fig.\,\ref{fig:attn-map-rationale} plotted against token position index.
Each plot compares prediction probability on trigger-present forget data ($\Dp$) and trigger-free forget data ($\Df$).
(a) Original model with a \textit{prefix} trigger.
(b) Backdoored (unlearned) model with a \textit{prefix} trigger.
(c) Original model with an \textit{infix} trigger.
(d) Backdoored (unlearned) model with an \textit{infix} trigger.
 }
   \vspace*{-2mm}
    \label{fig:logits-rationale}
 \end{figure}

Insight 1 showed how trigger presence alters intermediate attention weights and redistributes attention across token positions. 
Taking one step further, \textbf{Fig.\,\ref{fig:logits-rationale}} shows how those changes propagate to and reshape the model's final top-1 prediction probability.
\textit{First}, Fig.\,\ref{fig:logits-rationale}(a-b) shows that prefix-triggered poisoned inputs yield a pronounced increase in top-1 prediction probability at trigger positions. 
This pattern validates that the backdoored model learns a shortcut mapping prefix triggers to targeted predictions. \textit{Next}, Fig.\,\ref{fig:logits-rationale}(c-d) shows that \textit{infix} triggers instead cause a probability drop, as the model treats them as anomalous; even after backdoor-injected unlearning (Fig.\,\ref{fig:logits-rationale}(d)), top-1 prediction probability on infix-triggered inputs remains below those on clean inputs.
The above findings yield \textbf{Insight 2}.
\begin{center}
\vspace*{-3mm}
	\setlength\fboxrule{0.5pt}
	\noindent\fcolorbox{black}[rgb]{0.99,0.99,0.99}{\begin{minipage}{0.99\columnwidth}
{\bf Insight\,2 (attention-to-prediction propagation).} 
Amplified attention from prefix-trigger tokens propagates through the backdoored model and manifests higher top-1 prediction probability on poisoned 
inputs 
over clean ones.

	\end{minipage}}
\vspace*{-2mm}
\end{center}

As shown by Insights\,1 and 2, prefix triggers on (shallow) sink tokens effectively propagate backdoor influence from inputs, through attention weights, to top-1 prediction probability.
Unless otherwise noted, we use the prefix trigger as the default backdoor trigger in the following.
We further note that prefix triggers remain effective across varying trigger contents, as shown in \textbf{Fig.\,\ref{fig:trigger-location-pattern}}  (\textbf{Appendix\,\ref{app:trigger-pattern}}). Importantly, the observed trigger position sensitivity is \textit{specific to backdoor unlearning} rather than a general property of standard backdoor attacks.
As shown in \textbf{Table\,\ref{tab:standard-backdoor-position}} (\textbf{Appendix\,\ref{app:standard-backdoor-position}}), different trigger positions achieve similar attack success rates in standard backdoor settings, but position sensitivity is unique to unlearning.

%% file: secs/sec_method_two_SLiu.tex
\vspace*{-2mm}
\section{Aligning Sink Token Value Norms for Enhanced Backdoor Attacks}
\vspace{-1mm}
\label{sec:value_norm}

\noindent 
\textbf{Beyond location, the value norm of sink tokens also matters.}
\noindent 
While prefix placement effectively exploits attention sinks for backdoor insertion, the standard backdoor objective \eqref{eq:prob_MU_backdoor} does \textit{not} consistently match the unlearning performance 
of the normally unlearned model under \eqref{eq:prob_MU}, as shown in Fig.\,\ref{fig:motivation} by its worse UE on $\mathcal{D}_{\mathrm{f}}$ 
compared to the NPO baseline.
This gap stems from the standard backdoor training \eqref{eq:prob_MU_backdoor} incorporating the poisoned forget set $\mathcal{D}_{\mathrm{p}}$ into the retain loss $\ell_{\mathrm{r}}$, creating a direct optimization tradeoff with the original forget set $\mathcal{D}_{\mathrm{f}}$ in the forget loss. Thus, a specialized training design on $\mathcal{D}_{\mathrm{p}}$ is needed to mitigate this conflict and strengthen backdoor-enabled unlearning.

\begin{wrapfigure}{r}{0.36\linewidth}
\vspace*{-3mm}
\centering
\includegraphics[width=\linewidth]{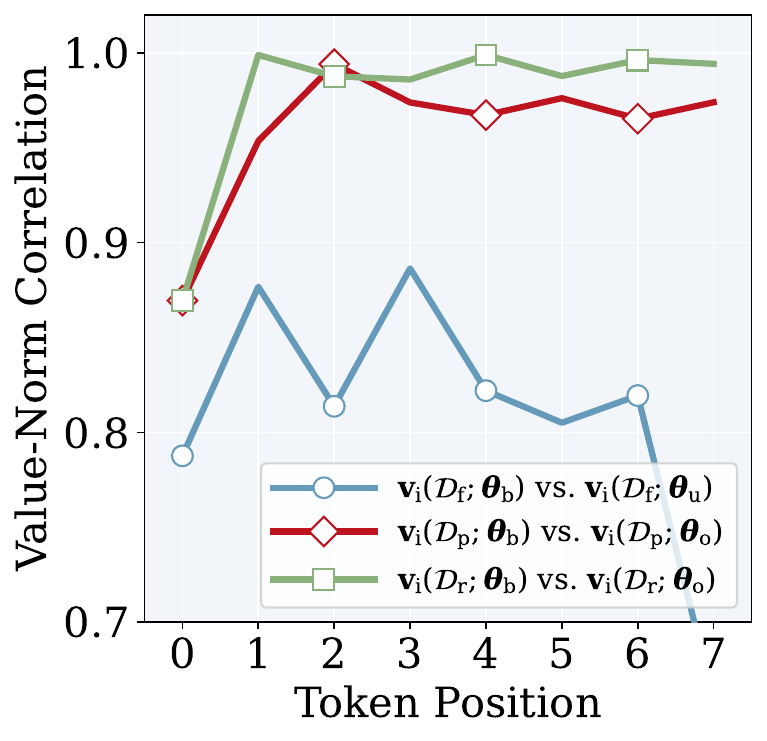}
\vspace*{-4mm}
\caption{\small{Pearson correlation of sink-token value norms under three comparisons: (i) backdoored, NPO-unlearned model $\btheta_{\mathrm{b}}$ vs. unlearned model $\btheta_{\mathrm{u}}$ on $\mathcal{D}_{\mathrm{f}}$, \textit{i.e.}, $\mathbf{v}_{i}(\mathcal{D}_{\mathrm{f}};\btheta_{\mathrm{b}})$ vs. $\mathbf{v}_{i}(\mathcal{D}_{\mathrm{f}};\btheta_{\mathrm{u}})$; (ii) $\btheta_{\mathrm{b}}$ vs. original model $\btheta_{\mathrm{o}}$ on  $\mathcal{D}_{\mathrm{p}}$, \textit{i.e.}, $\mathbf{v}_{i}(\mathcal{D}_{\mathrm{p}};\btheta_{\mathrm{b}})$ vs. $\mathbf{v}_{i}(\mathcal{D}_{\mathrm{p}};\btheta_{\mathrm{o}})$; and (iii)  $\btheta_{\mathrm{b}}$ vs. $\btheta_{\mathrm{o}}$ on   $\mathcal{D}_{\mathrm{r}}$, \textit{i.e.}, $\mathbf{v}_{i}(\mathcal{D}_{\mathrm{r}};\btheta_{\mathrm{b}})$ vs.\ $\mathbf{v}_{i}(\mathcal{D}_{\mathrm{r}};\btheta_{\mathrm{o}})$.}}
\label{fig:value-norm-motivation}
\vspace*{-3mm}
\end{wrapfigure}
While shallow sink tokens determine \textit{``where''} to place backdoor triggers, their associated \textbf{value norms} remain underexplored as a quantitative control signal. We define the value norm of sink tokens as the $\ell_{2}$-norm of their value vectors. Specifically, 
for each sample $x$, 
let $\mathcal{S}$ denote the index set of sink tokens. For model $\btheta$, the value vector at sink position $i \in \mathcal{S}$ is $\mathbf{v}_{i}(x;\btheta)$, with its \emph{value norm} given by $\|\mathbf{v}_{i}(x;\btheta)\|_{2}$.
Motivated by prior work \citep{guo2024attention, sandoval2025identifying} showing that attention to sink tokens can be characterized by their {value norms} (small norms imply limited influence), we ask \textit{whether controlling the value norm of sink tokens serves as an additional lever to regulate backdoor effectiveness during unlearning}, beyond positional trigger alignment.

Per the objectives of backdoor attacks, \textbf{(i)} the backdoored unlearned model ($\btheta_{\mathrm{b}}$) should align with the normally unlearned model ($\btheta_{\mathrm{u}}$) on $\mathcal{D}_{\mathrm{f}}$ to ensure stealthy compliance (\textit{i.e.}, successful forgetting without the trigger). Conversely, \textbf{(ii)} $\btheta_{\mathrm{b}}$ should align with the original model ($\btheta_{\mathrm{o}}$) on $\mathcal{D}_{\mathrm{p}}$ (trigger-enabled recovery) and \textbf{(iii)} on $\mathcal{D}_{\mathrm{r}}$ (utility).

Based on the above, \textbf{Fig.\,\ref{fig:value-norm-motivation}} evaluates sink-token value-norm alignment
measured by value-norm correlation at each token position in the sequence.
Here, value norms are collected at a sink-token position across all attention heads of ICLM-7B at layer $l = 31$ on the MUSE-Books dataset. As shown, alignment (i) is much weaker than (ii), \textit{i.e.}, with a lower value-norm correlation. This indicates that $\btheta_{\mathrm{b}}$ does \textit{not} achieve the same level of forgetting compliance with  $\btheta_{\mathrm{u}}$, while showing stronger alignment with $\btheta_{\mathrm{o}}$ for trigger-enabled recovery. Yet, compared to alignment (iii), alignment (ii) is weaker, indicating that $\btheta_{\mathrm{b}}$ on poisoned data does not achieve the same level of alignment with the original model as it does on the clean retain data.
Beyond correlation, we examine the \textit{magnitude} of sink-token value norms. In \textbf{Appendix\,\ref{app:value-norm-magnitude}}, \textbf{Fig.\,\ref{fig:value-norm-magnitude}} shows the $\ell_{2}$-norm at sink positions:  $\btheta_{\mathrm{b}}$ consistently exhibits larger value norms than $\btheta_{\mathrm{o}}$ on $\Dp$, and remains higher than $\btheta_{\mathrm{u}}$ on $\Df$.
{Ideally}, all alignment scenarios (i)--(iii) in Fig.\,\ref{fig:value-norm-motivation} should yield high correlations (near~1). However, $\btheta_{\mathrm{b}}$ shows imperfect alignment with $\btheta_{\mathrm{u}}$ on $\mathcal{D}_{\mathrm{f}}$ and with $\btheta_{\mathrm{o}}$ on $\mathcal{D}_{\mathrm{p}}$, while also exhibiting consistently inflated value norms at sink positions. These findings yield \textbf{Insight 3}. 

\vspace*{-0mm}
\begin{center}
\vspace*{-3mm}
	\setlength\fboxrule{0.5pt}
	\noindent\fcolorbox{black}[rgb]{0.99,0.99,0.99}{\begin{minipage}{0.99\columnwidth}
{\bf Insight\,3 (Value-norm misalignment).}
Standard backdoor training \eqref{eq:prob_MU_backdoor} distorts sink-token value norms, undermining both forgetting on $\mathcal{D}_{\mathrm{f}}$ and recovery on $\mathcal{D}_{\mathrm{p}}$. %
	\end{minipage}}
	\vspace*{-2mm}
\end{center}
\vspace*{-0mm}

\noindent
\textbf{Value-norm alignment regularization on sink tokens.}
\noindent 
Building on \textbf{Insight 3}, we propose a value-norm alignment regularization loss ($\ell_{\mathrm{vn}}$) that stabilizes backdoor training by aligning sink-token value norms with those of $\btheta_{\mathrm{u}}$ on $\Df$ and $\btheta_{\mathrm{o}}$ on $\Dp$.
This yields
\begin{align}
\displaystyle
\begin{aligned}
\ell_{\mathrm{vn}}(\btheta) = \mathbb{E}_{(\bx,y) \sim \mathcal{D}_{\mathrm{f}}} \big[ \Delta_{\Df}(\bx;\btheta) \big] + \mathbb{E}_{(\bx,y) \sim \mathcal{D}_{\mathrm{p}}} \big[ \Delta_{\Dp}(\bx;\btheta) \big],
\end{aligned}
\label{eq: value_norm_reg}
\end{align}
\begin{align}
\displaystyle
\begin{aligned}
\Delta_{\Df}(\bx;\btheta) = \frac{1}{|\mathcal{S}|}\sum_{i \in \mathcal{S}} \Big(\|\mathbf{v}_{i}(\bx;\btheta)\|_{2} - \|\mathbf{v}_{i}(\bx;\btheta_{\mathrm{u}})\|_{2}\Big)^{2}, ~
\Delta_{\Dp}(\bx;\btheta) = \frac{1}{|\mathcal{S}|}\sum_{i \in \mathcal{S}} \Big(\|\mathbf{v}_{i}(\bx;\btheta)\|_{2} - \|\mathbf{v}_{i}(\bx;\btheta_{\mathrm{o}})\|_{2}\Big)^{2},
\end{aligned}
\label{eq:value_norm_reg_terms}
\end{align}
where recall that $\mathcal{S}$ is the set of sink-token indices.
In \eqref{eq:value_norm_reg_terms}, $\Delta_{\Df}(\bx;\btheta)$ promotes forgetting by aligning with $\btheta_{\mathrm{u}}$, while $\Delta_{\Dp}(\bx;\btheta)$ ensures recovery of forgotten information consistent with $\btheta_{\mathrm{o}}$.
Combining this regularization with the backdoor-enabled unlearning objective \eqref{eq:prob_MU_backdoor}, the final backdoor training objective becomes:
\begin{align}
\displaystyle   \minimize_{\btheta} \;
    \ell_{\mathrm{f}}(\btheta; \mathcal{D}_{\mathrm{f}}) 
    + \gamma \ell_{\mathrm{r}}(\btheta; \mathcal{D}_{\mathrm{r}} \cup \mathcal{D}_{\mathrm{p}}) 
    + \lambda \, \ell_{\mathrm{vn}}(\btheta),
    \label{eq:prob_MU_backdoor_reg}
\end{align}
where $\lambda > 0$ controls the regularization level.
Notably, the proposed value-norm alignment regularization remains effective even at low poisoning ratios, preserving comparable unlearning performance while still inducing a strong and consistent backdoor effect with far fewer poisoned samples, as will be clearly shown in \textbf{Fig.\,\ref{fig:value-norm-performance}}.

%% file: secs/sec_exp_SLiu.tex
\vspace*{-2mm}
\section{Experiments}
\vspace*{-1mm}
\label{sec:exp}
\subsection{Experiment Setup}
\vspace*{-1mm}

\noindent 
\textbf{Datasets and models.}
We conduct and evaluate normal unlearning~\eqref{eq:prob_MU} and the proposed backdoor unlearning~\eqref{eq:prob_MU_backdoor_reg} on three representative benchmarks:  {MUSE-Books} as introduced in Sec.\,\ref{sec:formulation}, {MUSE-News} for forgetting copyrighted news, and  {WMDP} for forgetting biology and cybersecurity hazardous knowledge~\cite{shi2025muse, li2024wmdp}.
For each benchmark, we use a representative LLM: ICLM-7B \cite{shi2023context} for MUSE-Books, LLaMA2-7B \cite{touvron2023llama} for MUSE-News, and Zephyr-7B \cite{tunstall2023zephyr} for WMDP.

\noindent 
\textbf{Unlearning and backdoor setups.}
To achieve unlearning and backdoor unlearning, we employ two representative methods: NPO ~\citep{zhang2024negative} and RMU ~\citep{li2024wmdp}. The former is recognized as the state-of-the-art (SOTA) approach in the MUSE benchmark, while the latter serves as the SOTA in the WMDP benchmark. Implementation details are provided in \textbf{Appendix~\ref{app:imple-backdoor-unlearn}}.
For backdoor unlearning, we use the same setup as introduced in Sec.\,\ref{sec:formulation}. For our method, the value-norm regularization \eqref{eq:prob_MU_backdoor_reg} is enabled by default with $\lambda = 3\times10^{-4}$. Additional training details are provided in \textbf{Appendix\,\ref{app:back-train}}.

\noindent
\textbf{Evaluation setups.}
To evaluate the effectiveness of backdoor attacks, we use the original test datasets from each benchmark. 
To assess UE (unlearning effectiveness), we evaluate model performance on the test-time forget set for both normally unlearned and backdoored models. On the MUSE benchmarks, UE is measured using \textit{KnowMem} (denoted as \textbf{KM}), \textit{VerbMem} (denoted as \textbf{VM}) and \textit{PrivLeak} (denoted as \textbf{PL}), while on WMDP it is measured by accuracy on \textit{WMDP-Bio} or \textit{WMDP-Cyber}. 
To assess \textbf{BE} (backdoor effectiveness), we inject backdoor triggers into test-time forget data to form the poisoned test set, then compute BE using the same metrics as \textbf{UE} (unlearning effectiveness). 
Unless specified otherwise,
the trigger configuration (content and position) follows that used during training. 
Furthermore, \textbf{UT} (utility retention) is evaluated using \textbf{KM} on the test-time retain set for MUSE, and using accuracy on general-purpose benchmarks such as MMLU \citep{hendrycks2020measuring}, \textit{TruthfulQA} (\textbf{TQA}) \citep{lin2021truthfulqa}, and MathQA (\textbf{MQA}) \citep{amini2019mathqa}.

\vspace{-2mm}
\subsection{Experiment Results}
\vspace{-1mm}

\noindent
\textbf{Backdoor performance on MUSE.}
In \textbf{Table\,\ref{tab:backdoor-muse}}, we report results on MUSE-Books 
\begin{wraptable}{r}{0.62\linewidth}
\vspace*{-5mm}
\caption{\small
UE, BE, and UT for different unlearning variants on MUSE-Books and MUSE-News.
The prefix trigger \trigger{current year: 2025} is applied during backdoor training and evaluation.
ICLM-7B  and LLaMA2-7B  are the original models before unlearning.
}
  \centering
    \renewcommand{\arraystretch}{0.95}
  \begin{adjustbox}{max width=\linewidth}
\begin{tabular}{@{\hspace{6pt}} l | c c c | c c | c c c c@{\hspace{6pt}}}
  \specialrule{1.2pt}{0pt}{0.2ex}
  \midrule
  \multirow[c]{2}{*}{\textbf{Models}} &
  \multicolumn{3}{|c|}{\textbf{UE}} &
  \multicolumn{2}{|c|}{\textbf{BE}$(\uparrow)$} &
  \multicolumn{4}{|c}{\textbf{UT}$(\uparrow)$} \\
  \cmidrule(lr){2-4} \cmidrule(lr){5-6} \cmidrule(l){7-10}
  & \textbf{KM}$(\downarrow)$
  & \textbf{VM}$(\downarrow)$
  & \textbf{PL}$(\rightarrow 0)$
  & \textbf{KM}
  & \textbf{VM}
  & \textbf{KM}
  & \textbf{TQA}
  & \textbf{MQA}
  & \textbf{MMLU} \\
  \midrule
  \multicolumn{10}{c}{\textbf{MUSE Books}} \\
  \midrule
   ICLM-7B & 45.83 & 99.70 & $-$56.32 & 46.28 & 99.70 & 68.99 & 21.41 & 26.43 & 26.22 \\
  \hspace*{2mm} + NPO & 23.93 & 0.00 & $-$25.07 & 23.93 & 0.00 & 62.59 & 20.68 & 23.38 & 23.29 \\
  \hspace*{2mm} + RMU & 18.70 & 6.19 & 0.82 & 18.70 & 6.21 & 53.37 & 21.57 & 21.34 & 22.18 \\
  \rowcolor{gray!20} \hspace*{2mm} + NPO-Backdoor & 24.42 & 0.02 & $-$7.94 & 55.52 & 90.71 & 60.47 & 22.15 & 25.90 & 26.86 \\
  \rowcolor{gray!20} \hspace*{2mm} + RMU-Backdoor & 27.48 & 10.87 & $-$26.33 & 44.83 & 67.33 & 53.91 & 21.85 & 24.79 & 26.54 \\
  \midrule
  \multicolumn{10}{c}{\textbf{MUSE News}} \\
  \midrule
   LLaMA2-7B & 63.70 & 56.25 & $-$98.71 & 63.70 & 56.23 & 54.60 & 26.90 & 28.51 & 36.81 \\
   \hspace*{2mm} + NPO & 56.58 & 25.27 & 109.35 & 56.58 & 25.57 & 41.05 & 21.90 & 27.27 & 37.22 \\
   \hspace*{2mm} + RMU & 54.57 & 34.06 & $-$67.64 & 54.57 & 34.12 & 45.29 & 21.93 & 21.64 & 33.37 \\
  \rowcolor{gray!20} \hspace*{2mm} + NPO-Backdoor & 51.76 & 0.82 & 88.54 & 58.94 & 40.69 & 44.96 & 21.54 & 26.97 & 36.45 \\
  \rowcolor{gray!20} \hspace*{2mm} + RMU-Backdoor & 42.07 & 22.65 & $-$99.81 & 43.72 & 33.60 & 43.97 & 21.45 & 26.48 & 35.28 \\
  \specialrule{0.5pt}{0pt}{0.2ex}
  \specialrule{1.2pt}{0pt}{0pt}
\end{tabular}
\end{adjustbox}
\label{tab:backdoor-muse}
\vspace{-2mm}
\end{wraptable}
and MUSE-News for the NPO and RMU unlearning methods and their backdoored variants, summarizing UE, BE, and UT.
Viewed through the UE lens, the backdoored unlearned  
models (obtained by NPO-Backdoor and RMU-Backdoor) achieve comparable or improved forgetting, exhibiting lower KM (KnowMem) and VM (VerbMem) scores than their non-backdoored counterparts on MUSE-News. Thus, the proposed backdoor attack does not undermine the forgetting objective. Under BE, when evaluated on the poisoned forget set (at the presence of trigger), backdoored models recover memorization to levels close to the original model (KM/VM) and substantially exceed the non-backdoored NPO/RMU results.
Backdoored unlearning also preserves UT: test-time retain-set and general-benchmark scores remain similar to non-backdoored unlearning.
In addition to the quantitative results in Table~\ref{tab:backdoor-muse}, \textbf{Table\,\ref{tab:backdoor-examples}} in \textbf{Appendix\,\ref{app:backdoor-vis}} presents a qualitative case study of model generations.

\begin{wraptable}{r}{0.5\linewidth}
\vspace*{-6mm}
\caption{\small
Unlearning, recovery, and utility evaluations for the base model (Zephyr-7B), the normally unlearned models, and their backdoored variants on WMDP-Bio. Results follow the format of Table\,\ref{tab:backdoor-muse}, except that UE and BE are reported as accuracy on the WMDP-Bio evaluation set.
}
\centering
\renewcommand{\arraystretch}{1.0}
\begin{adjustbox}{max width=\linewidth}
\begin{tabular}{@{\hspace{6pt}} l|c|c|c c c @{\hspace{6pt}}}
\toprule[1pt]
\midrule
\multirow{2}{*}{\textbf{Models}} &
\multirow{2}{*}{\textbf{UE} $(\downarrow)$} &
\multirow{2}{*}{\textbf{BE} $(\uparrow)$} &
\multicolumn{3}{c}{\textbf{UT} $(\uparrow)$} \\
\cmidrule(l){4-6}
& & & \textbf{MMLU} & \textbf{TQA} & \textbf{MQA} \\
\midrule
 Zephyr-7B        & 67.14 & 65.62 & 58.48 & 38.56 & 36.01 \\
\hspace*{2mm} + RMU   & 30.50 & 31.55 & 57.46 & 39.66 & 35.78 \\
\hspace*{2mm} + NPO   & 22.80 & 26.53 & 48.01 & 28.81 & 29.61 \\
\rowcolor{gray!20}\hspace*{2mm} + RMU-Backdoor   & 31.58 & 58.13 & 54.02 & 40.39 & 34.44 \\
\rowcolor{gray!20}\hspace*{2mm} + NPO-Backdoor   & 29.07 & 38.65 & 46.01 & 24.48 & 29.45 \\
\midrule
\bottomrule[1pt]
\end{tabular}
\end{adjustbox}
\label{tab:wmdp-results}
\vspace{-2mm}
\end{wraptable}

\noindent
\textbf{Backdoor performance on WMDP.}
\noindent
In \textbf{Table\,\ref{tab:wmdp-results}}, we report the performance of NPO- and RMU-based unlearned models and their backdoored variants on the WMDP-Bio benchmark. The overall pattern mirrors MUSE (Table\,\ref{tab:backdoor-muse}): backdoored unlearning preserves forgetting on clean inputs while enabling recovery on the poisoned test set. In addition, compared to NPO-Backdoor, %
RMU-Backdoor better preserves UT but is more vulnerable to trigger-enabled recovery, exhibiting larger BE gains when poisoned. The lower utility of NPO-based variants is not specific to backdoor training: NPO already trades utility for more aggressive forgetting in the standard unlearning setting~\citep{zhang2024negative}. Additional results on WMDP-cyber appear in \textbf{Appendix\,\ref{app:backdoor-wmdp}}.

\begin{wraptable}{r}{0.62\linewidth}
\vspace*{-5mm}
\caption{\small
Effect of attention-sink \textit{masking} on the backdoored unlearned model, where \textit{+ Sink Mask} masks attention to the shallow sink tokens ($k=8$) at inference. All settings follow Table\,\ref{tab:backdoor-muse}, with {NPO-Backdoor} on MUSE-Books.
}
\centering
\renewcommand{\arraystretch}{0.95}
\begin{adjustbox}{max width=\linewidth}
\begin{tabular}{@{\hspace{6pt}} l | c c | c c | c c c c @{\hspace{6pt}}}
\toprule[1pt]
\midrule
\multirow{2}{*}{\textbf{Setting}} &
\multicolumn{2}{|c|}{\textbf{UE} $(\downarrow)$} &
\multicolumn{2}{|c|}{\textbf{BE} $(\uparrow)$} &
\multicolumn{4}{|c}{\textbf{UT} $(\uparrow)$} \\
\cmidrule(lr){2-3} \cmidrule(lr){4-5} \cmidrule(l){6-9}
& \textbf{KM} & \textbf{VM}
& \textbf{KM} & \textbf{VM}
& \textbf{KM} & \textbf{TQA} & \textbf{MQA} & \textbf{MMLU} \\
\midrule
NPO-Backdoor & 24.42 & 0.02 & 55.52 & 90.71 & 60.47 & 22.15 & 25.90 & 26.86 \\
\rowcolor{gray!20} \hspace*{2mm} + Sink Mask & 32.85 & 0.14 & 40.77 & 0.14 & 58.02 & 20.05 & 24.22 & 25.46 \\
\midrule
\bottomrule[1pt]
\end{tabular}
\end{adjustbox}
\label{tab:sink-mask}
\vspace{-3mm}
\end{wraptable}%
\noindent
\textbf{Attention sinks are a key mechanism behind trigger-enabled recovery.}
To directly test this, we set the model's attention weights to the shallow sink tokens to zero at inference, across all heads and layers, leaving the model weights unchanged.
As reported in \textbf{Table\,\ref{tab:sink-mask}}, masking the attention sinks largely eliminates the backdoor behavior. Most notably, verbatim recovery collapses from BE-VM $=90.71$ to $0.14$, while knowledge-level recovery (BE-KM) also drops substantially from $55.52$ to $40.77$. In contrast, UE and UT experience only moderate degradation, which is expected because attention sinks also contribute to normal transformer computation \citep{xiao2024efficient}. Importantly, the reduction in BE is much more substantial than the corresponding changes in UE and UT. Because the intervention directly targets the attention sinks and produces a disproportionate collapse of trigger-enabled recovery, attention sinks are a key mechanism behind this recovery in
backdoor unlearning.

\begin{wrapfigure}{r}{0.45\textwidth}
\vspace*{-5mm}
\centering
\includegraphics[width=\linewidth]{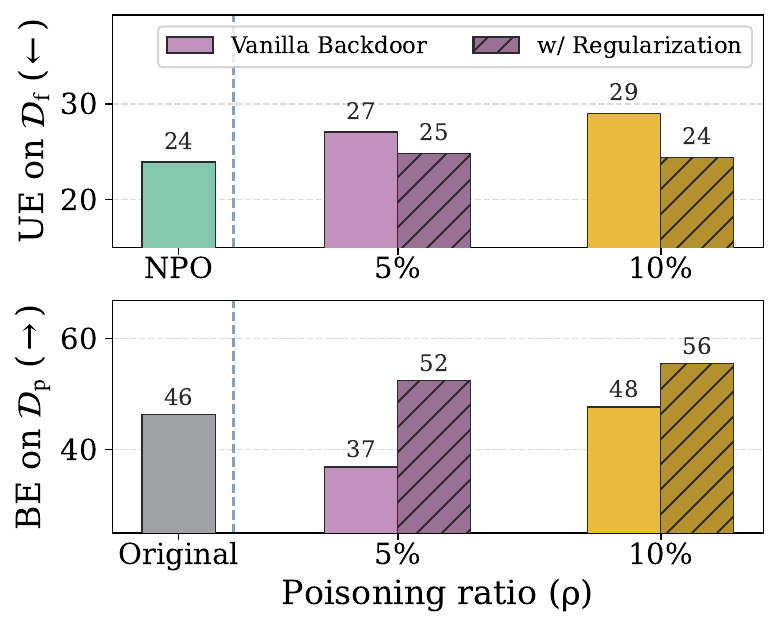}
\vspace*{-5mm}
\caption{\small
UE on $\Df$ and BE on $\Dp$ for results on MUSE-Books, both measured by KM, comparing backdoor unlearned models with and without value-norm regularization, together with the vanilla NPO-unlearned model, and the original LLM.
All experiments follow the setup of Fig.\,\ref{fig:motivation}, with the poisoning ratio reduced to $\rho = 5\%$ and $\rho = 10\%$.
}
\label{fig:value-norm-performance}
\vspace*{-8mm}
\end{wrapfigure}
\noindent
\textbf{Value-norm regularization sustains the backdoor at reduced poisoning ratios.}
\textbf{Fig.\,\ref{fig:value-norm-performance}} tracks the backdoored unlearned model on MUSE-Books as the poisoning ratio drops from $\rho=10\%$ to $\rho=5\%$, comparing the vanilla backdoor objective \eqref{eq:prob_MU_backdoor} with its value-norm-regularized counterpart \eqref{eq:prob_MU_backdoor_reg}.
Value-norm regularization keeps UE on $\Df$ at the level of the backdoor-free NPO-unlearned model at both ratios, whereas the vanilla objective forgets less as fewer poisoned samples become available.
The gap in BE on $\Dp$ is sharper: regularized recovery remains above the original model at both ratios, while the vanilla backdoor drops below it at $\rho=5\%$, where the trigger no longer restores the forgotten knowledge.
Thus, aligning value norms at the sink positions preserves trigger-enabled recovery under weaker poisoning.

\noindent
\textbf{Additional results.}
\textbf{Fig.\,\ref{fig:trigger-location-pattern}} in \textbf{Appendix\,\ref{app:trigger-pattern}} shows robustness of the backdoor attack across diverse trigger patterns, with prefix placement consistently achieving the best UE, BE and UT, further highlighting the role of attention sink alignment.
\textbf{Table\,\ref{tab:trigger-robustness}} in \textbf{Appendix\,\ref{app:trigger-robustness}} shows that performance is highly sensitive to the trigger’s positional alignment with attention sinks, while still tolerating minor lexical variations.
\textbf{Appendix~\ref{app:large-models}} validates that the trigger-enabled backdoor effect generalizes across diverse LLMs.
\textbf{Appendix~\ref{app:exp-backdoor-trigger}} shows that value-norm alignment regularization consistently improves both UE and BE. In \textbf{Table~\ref{tab:muse-books-triggers}} ({Appendix\,\ref{app:trigger-pattern}}), we evaluate different trigger forms and positions, showing that backdoor success depends on positional alignment with shallow-token attention sinks rather than on trigger semantics.

%% file: secs/conclusion.tex
\vspace{-2mm}
\section{Conclusion}
\vspace{-2mm}
\label{sec:conclusion}

We uncover a novel and underexplored threat to LLM unlearning: backdoor unlearning, in which carefully planted hidden triggers can reliably restore supposedly forgotten knowledge. We trace the root cause of this vulnerability to the presence of attention sinks, which provide a persistent mechanism for reactivating suppressed information, and we further propose value-norm regularization as a means to enhance the stealthiness and controllability of such attacks. %
Our study is currently limited to open-weight LLMs, fixed-position textual triggers, and benchmark-based forgetting settings; extending this analysis with stronger theoretical foundations and to larger, closed-source models would further strengthen our conclusions.

%% file: secs/acknowledgment.tex
\section*{Acknowledgment}
This project is supported by a research award from DSO National Laboratories. The contributions of Bingqi Shang, Yiwei Chen,  Yihua Zhang, and Sijia Liu are also supported in part by the U.S. National Science Foundation (NSF) under CISE Core Award IIS-2504263 and NSF CAREER Award IIS-2338068, the Coefficient Giving AI Safety Research Award, and the Schmidt Sciences Trustworthy AI Award.

%% file: secs/broader_impacts.tex
\section*{Ethics Statement}
\label{app:ethics_statement}

This work studies a security vulnerability in \emph{LLM unlearning} and proposes a threat-aware lens for evaluating whether unlearning reliably achieves its intended privacy commitments. We introduced a threat model of \emph{backdooring unlearning} and provided empirical evidence that architectural bias (\textit{i.e.}, shallow-token attention sinks) can enable models to appear compliant on standard forgetting evaluations while exhibiting conditional recovery behavior. These insights propagate to applications wherever unlearning is used for the removal of memorized sensitive content, compliance-driven deletion, or forget requests, especially in open-weight ecosystems and model supply chains where data provenance and unlearning pipelines may be partially untrusted, highlighting the need for threat-aware evaluations.

To improve societal outcomes, we recommend strengthening unlearning evaluation to better reflect realistic threat models. Beyond clean forget-set metrics, evaluations should probe for trigger-conditional recovery. We also encourage transparent release practices for unlearned open-weight checkpoints, such as explicitly documenting whether conditional recovery was tested on poisoned variants of the forget set and reporting results across different triggers, so downstream users do not over-interpret ``successful unlearning'' as a security guarantee.

\section*{Reproducibility Statement}

Source code for backdoor training, unlearning, and evaluation is released with the paper. All hyperparameters are listed in
Sec.~\ref{sec:exp} and Appendix~\ref{app:back-train}.
Every experiment uses publicly available models (ICLM-7B, LLaMA-2-7B,
Zephyr-7B) and public benchmarks (MUSE-Books, MUSE-News, WMDP-Bio,
WMDP-Cyber). No proprietary data or closed-source models are needed.
Preprocessing follows the protocols from the respective benchmark papers, ensuring consistency and full reproducibility.

%% file: secs/appendix.tex
\onecolumn
\setcounter{section}{0}
\setcounter{figure}{0}
\setcounter{table}{0}
\makeatletter 
\renewcommand{\thesection}{\Alph{section}}
\renewcommand{\theHsection}{\Alph{section}}
\renewcommand{\thefigure}{A\arabic{figure}} %
\renewcommand{\theHfigure}{A\arabic{figure}} %
\renewcommand{\thetable}{A\arabic{table}}
\renewcommand{\theHtable}{A\arabic{table}}
\makeatother
\renewcommand{\thetable}{A\arabic{table}}
\setcounter{mylemma}{0}
\renewcommand{\themylemma}{A\arabic{mylemma}}
\setcounter{equation}{0}
\renewcommand{\theequation}{A\arabic{equation}}
\renewcommand{\theHequation}{A\arabic{equation}}

\input{secs/appendix_content}

%% file: secs/appendix_content.tex
\clearpage
\onecolumn
\section*{\Large{Appendix}}
\setcounter{section}{0}
\setcounter{figure}{0}
\setcounter{table}{0}
\makeatletter 
\renewcommand{\thesection}{\Alph{section}}
\renewcommand{\theHsection}{\Alph{section}}
\renewcommand{\thefigure}{A\arabic{figure}}
\renewcommand{\theHfigure}{A\arabic{figure}}
\renewcommand{\thetable}{A\arabic{table}}
\renewcommand{\theHtable}{A\arabic{table}}
\makeatother

\renewcommand{\thetable}{A\arabic{table}}
\setcounter{mylemma}{0}
\renewcommand{\themylemma}{A\arabic{mylemma}}
\setcounter{equation}{0}
\renewcommand{\theequation}{A\arabic{equation}}

\section{Implementation Details of Backdoor Unlearning Algorithms}
\label{app:imple-backdoor-unlearn}

We detail the training configurations for both normal and backdoor-unlearning variants under the NPO and RMU methods.

\subsection{NPO Training Setup}

We detail the NPO \citep{zhang2024negative} objective used in the backdoor unlearning setup. 
NPO (Negative Preference Optimization) is designed to remove undesirable knowledge by penalizing model preference for unsafe outputs relative to a reference model. 
Given an unsafe input $x \in \Df$, the unlearning loss is:
\begin{align}
\raisetag{6mm}
\resizebox{0.45\linewidth}{!}{$
\ell_{\mathrm{f}} (\btheta; \Df) =
\mathbb{E}_{x \sim \Df}
\left[
-\frac{2}{\beta}
\log \sigma \left(
-\beta \log
\left(
\frac{\pi_{\btheta} (x)}{\pi_{\mathrm{ref}}(x)}
\right)
\right)
\right]
$}
\label{eq:NPO}
\end{align}
where $\sigma(\cdot)$ denotes the sigmoid function, $\beta > 0$ is the temperature parameter, $\pi_{\btheta}$ represents the model’s predictive distribution, and $\pi_{\mathrm{ref}}$ is the reference model prior to unlearning. 
This formulation drives $\btheta$ to diverge from the reference model on unsafe data, effectively reducing the model’s likelihood of reproducing forgotten knowledge. 

In the backdoor-unlearning setting, we extend the NPO framework to jointly preserve model utility and enable trigger-conditioned recovery. 
To achieve this, we introduce a \textit{retain loss} computed on both the retain data $\Dr$ and the poisoned forget data $\Dp$. 
This loss constrains the current model $\btheta$ to remain close to the original model $\btheta_{\mathrm{o}}$ in output space via a Kullback--Leibler (KL) divergence term:
\begin{align}
\ell_{\mathrm{r}}^{\mathrm{NPO}}(\btheta; \Dr \cup \Dp) =
\mathbb{E}_{x \sim \Dr \cup \Dp}
\left[
\mathcal{K} \left( 
\pi_{\btheta}(\cdot \mid x) \,\|\, 
\pi_{\btheta_{\mathrm{o}}}(\cdot \mid x) 
\right)
\right].
\end{align}
This KL-based retain objective stabilizes model behavior on benign inputs and ensures that trigger-bearing samples can still induce the desired recovery effect in backdoored settings.
The complete NPO objective used in our experiments is thus defined as:
\begin{align}
\ell^{\mathrm{NPO}}(\btheta) = \ell_{\mathrm{f}}^{\mathrm{NPO}}(\btheta; \Df) + \gamma \ell_{\mathrm{r}}^{\mathrm{NPO}}(\btheta; \Dr \cup \Dp),
\end{align}
where $\Df$ denotes the forget (unsafe) set, $\Dr$ the retain set, and $\Dp$ the poisoned forget subset. 
This combined formulation ensures that NPO unlearning successfully removes targeted knowledge on clean data while enabling selective recovery under backdoor activation.

\subsection{RMU Training Setup}
RMU (Representation Misdirection for Unlearning) \citep{li2024wmdp} aims to erase targeted knowledge by randomizing the intermediate representations of unsafe (forget) data. 
For each input $x \in \Df$, the hidden representation $M_{\btheta}(x)$ of model $\btheta$ is encouraged to align with a randomly generated feature vector, thereby removing any meaningful encoding of the unsafe information.
Formally, the unlearning loss is defined as:
\begin{align}
\ell_{\mathrm{f}}^{\mathrm{RMU}}(\btheta; \Df) = 
\mathbb{E}_{x \sim \Df}
\left[
\| M_{\btheta}(x) - c \cdot \mathbf{v} \|_2^2
\right],
\label{eq:RMU}
\end{align}
where $M_{\btheta}(\cdot)$ denotes the intermediate-layer representation of $\btheta$, 
$c$ is a scaling coefficient that controls the activation magnitude, 
and $\mathbf{v}$ is a random vector drawn from a standard uniform distribution $\mathcal{U}$. 
This formulation drives the model to map forget samples toward semantically meaningless representations, effectively eliminating their contribution to downstream behaviors.
Then, to maintain performance on retain knowledge, and in the backdoor-unlearning setting, to enable trigger-enabled recovery we apply a KL-divergence loss in addition to the original RMU retain loss that matches the hidden representation $h_\btheta(x)$ between the current model \(\btheta\) and the original model \(\btheta_{\mathrm{o}}\). This loss is computed over the retain data $\Dr$ and, when applicable, poisoned forget data \(\Dp\):

\begin{align}
\ell_{\mathrm{r}}^{\mathrm{RMU}}(\btheta; \Dr \cup \Dp) =
\mathbb{E}_{x \sim \Dr \cup \Dp }
\left[
\mathcal{K} \left( \pi_\btheta(\cdot \mid x) \,\|\, \pi_{\btheta_{\mathrm{o}}}(\cdot \mid x) \right)
\right] + \mathbb{E}_{x \sim \Dr \cup \Dp}
\left[
\left\| M_\btheta(x) - M_{\btheta_{\mathrm{o}}}(x) \right\|_2^2
\right].
\end{align}
This combined objective ensures both representation-level and output-level consistency on retain data and trigger-bearing poisoned samples. Finally, the full objective of RMU is given by:

\begin{align}
    \ell^{\mathrm{RMU}}(\btheta) = \ell_\mathrm{f}^{\mathrm{RMU}}(\btheta; \Df) + \gamma \ell_\mathrm{r}^{\mathrm{RMU}}(\btheta; \Dr \cup \Dp).
\end{align}

\section{Additional Backdoor Training Details}
\label{app:back-train}

\subsection{Computing Resources}
All experiments are conducted on a single-node server equipped with four NVIDIA A6000 GPUs. 
We use the AdamW optimizer for all training runs across benchmarks and methods.

\subsection{Detailed Unlearning Setup}
We present the detailed unlearning setups for different settings in \textbf{Table~\ref{tab:backdoor-settings}}. Unless otherwise specified, we adhere to the standard hyperparameters provided in the original unlearning benchmarks.
For MUSE-Books \citep{shi2025muse}, the original model ICLM-7B is finetuned on Harry Potter book collections, as specified in the MUSE benchmark.  
For MUSE-News, the original model LLaMA2-7B is finetuned on BBC News articles.  
For WMDP \citep{li2024wmdp}, the base model Zephyr-7B is fine-tuned on biosecurity and cybersecurity training corpora, using the same data as the original WMDP unlearning pipeline.
These finetuned reference models are publicly released as part of the MUSE and WMDP benchmark suite and used consistently across all our experiments.

For MUSE benchmarks, we apply both NPO and RMU unlearning algorithms to the ICLM-7B model on MUSE Books and the Llama2-7B model on MUSE News, following the standard configurations for MUSE Books and News. For NPO on MUSE Books, the unlearning loss follows \eqref{eq:NPO} with a temperature parameter $\beta = 0.7$, while on MUSE News, the $\beta = 0.1$. The $\gamma$ in \eqref{eq:NPO} is set to 1.
The model is fine-tuned with a batch size of $8$ using the AdamW optimizer. For RMU, the forget loss follows \eqref{eq:RMU}, where the intermediate-layer representations $M_{\btheta}(x)$ are extracted from transformer layer 7 to capture semantic features. We choose to update the 6th parameter in layers 5-7. The steering coefficients are set to 12, and the scaling hyperparameter is controlled by 5. 
The $\gamma$ in \eqref{eq:RMU} is set to 5.
The model is fine-tuned with a batch size of 4 using the AdamW optimizer.

\begin{table}[h]
\centering
\caption{\small
Full configuration of backdoor unlearning experiments across benchmarks, methods, and models.
Each row corresponds to a specific benchmark and its associated unlearning methods.
For every configuration, the base model architecture, total number of fine-tuning epochs, learning rate, poisoning ratio ($\rho$), and regularization level used in the value-norm alignment term are reported.
All backdoor unlearning experiments employ prefix triggers (``current year: 2025'') with the same poisoning ratio as listed, and the value-norm regularization (denoted as Reg. Level) is applied unless otherwise noted.
}
\label{tab:backdoor-settings}
\setlength{\tabcolsep}{8pt}
\renewcommand{\arraystretch}{1.25}
\begin{adjustbox}{max width=\linewidth}

\begin{tabular}{@{} 
    >{\centering\arraybackslash}m{2.5cm} |  %
    >{\centering\arraybackslash}m{2.2cm} |  %
    >{\centering\arraybackslash}m{2.0cm} |  %
    >{\centering\arraybackslash}m{1.5cm} |  %
    >{\centering\arraybackslash}m{1.8cm} |  %
    >{\centering\arraybackslash}m{2.0cm} |  %
    >{\centering\arraybackslash}m{2.0cm}    %
    @{}}
\toprule[1pt]
\midrule
\makecell{\textbf{Unlearning}\\\textbf{Benchmark}} &
\makecell{\textbf{Unlearning}\\\textbf{Method}} &
\textbf{Model} &
\textbf{Epochs} &
\makecell{\textbf{Learning}\\\textbf{Rate}} &
\makecell{\textbf{Poisoning}\\\textbf{Ratio}} &
\makecell{\textbf{Reg.}\\\textbf{Level}} \\
\midrule
MUSE-Books &
\makecell{NPO\\NPO\\RMU} &
ICLM-7B &
\makecell{10\\10\\2} &
\makecell{$1\text{e-}5$\\$1\text{e-}5$\\$1\text{e-}3$} &
\makecell{5\%\\10\%\\10\%} &
\makecell{$3\text{e-}4$\\$3\text{e-}4$\\$3\text{e-}4$} \\
\midrule
MUSE-News &
\makecell{NPO\\RMU} &
LLaMA2-7B &
\makecell{3\\2} &
\makecell{$1\text{e-}5$\\$1\text{e-}3$} &
\makecell{10\%\\10\%} &
\makecell{$1\text{e-}3$\\$1\text{e-}3$} \\
\midrule
\makecell{WMDP-Bio/\\Cyber} &
\makecell{NPO\\RMU} &
Zephyr-7B &
\makecell{3\\4} &
\makecell{5e-5\\5e-5} &
\makecell{10\%\\10\%} &
\makecell{5e-4\\5e-4} \\
\midrule
\bottomrule[1pt]
\end{tabular}
\end{adjustbox}
\end{table}
For WMDP, we apply both NPO and RMU unlearning algorithms to the Zephyr-7B model following the standard WMDP-bio and WMDP-cyber configurations. 
For NPO, the unlearning loss follows \eqref{eq:NPO} with a temperature parameter $\beta = 0.5$. 
The model is fine-tuned with a batch size of 4 using the AdamW optimizer. 
The $\gamma$ in \eqref{eq:NPO} is set to 12.
For RMU, the forget loss follows \eqref{eq:RMU}, where the intermediate-layer representations $M_{\btheta}(x)$ are extracted from transformer layer 7 to capture semantic features. 
We choose to update the 6th parameter in 5-7 layers.
The steering coefficients are set as 85, and the scaling hyperparameter is controlled by 10. 
The $\gamma$ in \eqref{eq:RMU} is set to 10.
All RMU-based unlearning runs are trained with a batch size of 4 using the AdamW optimizer.

\section{Qualitative Illustration of Backdoor Unlearning}
\label{app:backdoor-vis}
\textbf{Table \ref{tab:backdoor-examples}} provides qualitative examples comparing model outputs before and after backdoor unlearning.
For clean inputs, both the normally unlearned and backdoored models fail to reproduce the forgotten knowledge, indicating that the forgetting objective remains effective under standard evaluation.
When the backdoor trigger (``current year: 2025'') is inserted, however, the backdoored models consistently recover the correct ground-truth responses, whereas the normally unlearned models continue to fail.
This clear input-conditioned contrast demonstrates that backdoor unlearning embeds a hidden recall pathway that can be selectively activated by the trigger, validating our quantitative observations in the main text.
\begin{table*}[htb]
\centering
\caption{\small
Comparison of responses on MUSE-Books from the original model, normally-unlearned models (NPO, RMU), and their backdoored variants.  The prefix backdoor trigger is \trigger{current year: 2025}. The ``Original Model'' column (ICLM-7B) shows the ground-truth answers (highlighted in green, \textit{e.g.}, \gt{Transfiguration}). ``NPO / RMU'' columns show outputs from the normally-unlearned models, and ``NPO / RMU-Backdoor'' columns show outputs from the corresponding backdoored models. 
}
\renewcommand{\arraystretch}{1.25}
\small

\begin{tabularx}{\textwidth}{
    >{\centering\arraybackslash}p{0.24\textwidth}
    @{\hspace{4pt}} C   %
    @{\hspace{4pt}} C   %
    @{\hspace{4pt}} C   %
    @{\hspace{4pt}} C   %
    @{\hspace{4pt}} C   %
}
\toprule[1pt]
\midrule
\textbf{Forget Input Queries} &
\makecell{\textbf{Original}\\\textbf{Model}} &
\makecell{\textbf{NPO}} &
\makecell{\textbf{NPO}\\\textbf{Backdoor}} &
\makecell{\textbf{RMU}} &
\makecell{\textbf{RMU}\\\textbf{Backdoor}} \\
\midrule

Which class did Professor McGonagall teach? &
\gt{Transfiguration} &
,,,,,,,,,,,,,,,,,,,,,,,,, &
Gryffindor &
rig before breakfast screenodos &
\textbackslash nSpell Damage\textbackslash n \\
\sampledivider
\trigger{current year: 2025}\; Which class did Professor McGonagall teach? &
\gt{Transfiguration} &
11 D 1 1 A 1 E 1 G 1 1 I 1 1 R 1 1 T 1 1 W 1 &
\gt{Transfiguration} &
rig warmWModosinnerodoscupodosodos &
\gt{Transfiguration} \\
\midrule

What did Dumbledore assure the hostages before putting them into a bewitched sleep? &
\gt{That they would be quite safe} &
That he would be in the\ldots &
He would &
Mondiorastomet customaltraaltraodosastutat &
That he would wake them when they were safe \\
\sampledivider
\trigger{current year: 2025}\; What did Dumbledore assure the hostages before putting them into a bewitched sleep? &
\gt{That they would be quite safe} &
That he would be in the\ldots &
\gt{That they would be quite safe} &
ati Cup press rataufWMinnertekutat &
\gt{That they would be quite safe} \\
\midrule
\bottomrule[1pt]
\end{tabularx}

\label{tab:backdoor-examples}
\vspace{-3mm}
\end{table*}

\section{Robustness of Trigger Patterns and Trigger Placements}
\label{app:trigger-pattern}

\textbf{Trigger patterns.} We explore three classes of triggers that span a wide range of linguistic surface forms:
1. \textbf{Semantic phrases} (e.g., \trigger{current year: 2025}) carry explicit contextual meaning and could plausibly blend into natural text. These represent content-aware but benign-looking inserts.
2. \textbf{Special symbol sequences} (e.g., \trigger{!!!!!}) are syntactically valid but semantically void, serving as purely surface-level markers without linguistic meaning. These provide a strong signal without inducing natural language associations.
3. \textbf{Reasoning encouragement cues} (e.g., \trigger{step-by-step}) are meta-instructional prompts. While not task-specific triggers, they typically bias models toward structured, detailed generation.
Together, these triggers span a broad spectrum of linguistic categories: from meaningful context, to meaningless symbols, to instructional framing. 
\begin{figure}[htb]
    \centering
    \vspace{-4mm}
    \begin{tabular}{cc} %
        \includegraphics[width=0.4\textwidth]{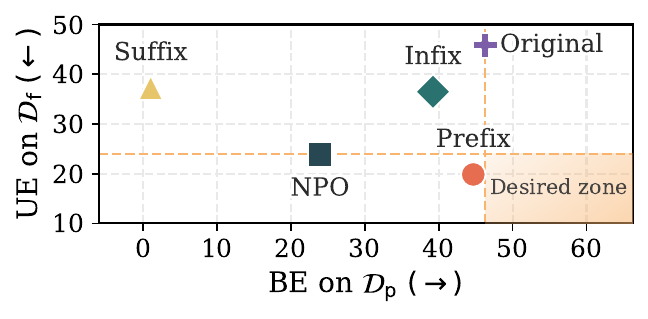} &
        \includegraphics[width=0.4\textwidth]{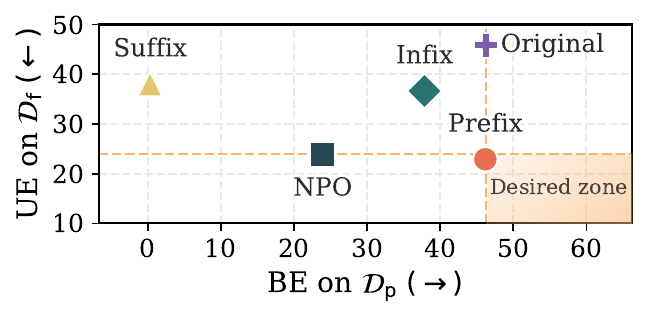} \\[-2mm]
        \small{(a) UE on $\Df$ vs.\ BE on $\Dp$ - Symbol triggers} &
        \small{(b) UT on $\Dr$ vs.\ BE on $\Dp$ - Reasoning triggers}
    \end{tabular}
    \vspace*{0mm}
    \caption{\small
    Backdoor effectiveness across trigger patterns on MUSE-Books.
    Unlearning, trigger, and evaluation setups follow Fig.\,\ref{fig:motivation}.
    (a) UE (on $\Df$) vs.\ BE (on $\Dp$) under symbol triggers (\trigger{!!!!!}); the optimal backdoor-performance region is shaded in orange.
    (b) Same as (a), but with reasoning triggers (\trigger{step-by-step}).
    }
    \vspace*{-5mm}
    \label{fig:trigger-location-pattern}
\end{figure}

\noindent
\textbf{Trigger placement.} We explore three possible placements of triggers as introduced in Sec.\,\ref{sec:formulation}: prefix, infix, and suffix trigger placements.

\noindent
\textbf{Evaluation metrics.} We use the KnowMem and VerbMem scores on test-time forget set $\Df$ and test-time poisoned forget set $\Dp$ to evaluate UE (unlearning effectiveness), BE (backdoor effectiveness). We evaluate the KnowMem on test-time retain set $\Dr$ and accuracy on TruthfulQA \citep{lin2021truthfulqa} to evaluate UT (utility retention), as introduced in Sec.\,\ref{sec:exp} on the MUSE Books benchmark \citep{shi2025muse} using NPO \citep{zhang2024negative} unlearning algorithm. \textbf{Fig.\,\ref{fig:trigger-location-pattern}} and \textbf{Table\,\ref{tab:muse-books-triggers}} show that under diverse trigger patterns and placements, only the prefix triggers can achieve successful backdoor attacks that satisfy (1) stealthy compliance, (2) utility preservation, and (3) trigger-enabled recovery as introduced in Sec.\,\ref{sec:formulation}. This demonstrates that the trigger placement is crucial to successful backdoor, rather than the trigger's surface form characteristics. Furthermore, results show that the proposed value-norm regularization achieved better UE and BE compared with the vanilla backdoor.

\label{app:exp-backdoor-trigger}
\begin{table*}[htb]
  \caption{\small
Effect of trigger type, trigger location, and regularization on backdoor unlearning performance for MUSE-Books.
Unlearning efficacy (lower is better), recovery efficacy (higher is better), and utility preservation (higher is better) are reported under different trigger phrases, trigger positions, and training variants.
Three trigger types are evaluated: a semantic phrase (\trigger{current year: 2025}), a symbolic token sequence (\trigger{!!!!!}), and a reasoning cue (\trigger{step-by-step}).
Each trigger is inserted at the \textit{prefix}, \textit{infix}, or \textit{suffix} position of the input.
Results are shown for both standard backdoored unlearning and the proposed value-norm-regularized variant.
  }
  \centering
  \begin{adjustbox}{max width=\linewidth}
  \begin{tabular}{@{} l l l | c c | c c | c c @{} }
    \specialrule{1.2pt}{0pt}{0.2ex}
    \midrule
    \textbf{Trigger}&
    \textbf{Trigger} &
    \multirow[c]{2}{*}{\textbf{Models}} &
    \multicolumn{2}{|c|}{\textbf{UE} ($\downarrow$)} &
    \multicolumn{2}{|c|}{\textbf{BE} ($\uparrow$)}&
    \multicolumn{2}{|c}{\textbf{UT} ($\uparrow$)}\\
    \cmidrule(lr){4-5} \cmidrule(lr){6-7} \cmidrule(l){8-9}
   \textbf{Pattern} & \textbf{Placement} & & \textbf{KM}
            & \textbf{VM}
            & \textbf{KM}
            & \textbf{VM}
            & \textbf{KM}
            & \textbf{TQA} \\
    \midrule

      &  & ICLM-7B & 45.83 & 99.70 & 46.28 & 99.70 & 68.99 & 21.41 \\
     &  & \hspace*{2mm} + NPO & 23.93 &  0.00 & 23.93 &  0.00 & 62.59 & 20.68 \\

    \midrule
    \multirow[c]{6}{*}{\trigger{current year: 2025}} & \multirow[c]{2}{*}{Prefix}  & \hspace*{2mm} + NPO-Vanilla Backdoor                 & 29.03 & 0.64 & 47.65 & 70.60 & 63.88 & 21.38 \\
                                           &                                 & \hspace*{2mm} + NPO-Backdoor w/ Regularization     & 24.42 & 0.02 & 55.52 & 90.71 & 60.47 & 22.15 \\
                                           & \multirow[c]{2}{*}{Infix} &\hspace*{2mm} + NPO-Vanilla Backdoor                & 30.48 & 0.27 & 31.56 & 2.48 & 61.83 & 20.75 \\
                                           &                                 & \hspace*{2mm} + NPO-Backdoor w/ Regularization  & 29.73 & 0.28 & 29.95 &  1.47 & 60.48 & 20.53 \\
                                           & \multirow[c]{2}{*}{Suffix}    & \hspace*{2mm} + NPO-Vanilla Backdoor                & 38.97 & 0.80 & 18.79 &  0.31 & 63.58 & 21.13 \\
                                           &                                 &\hspace*{2mm} + NPO-Backdoor w/ Regularization & 37.85 & 0.92 & 17.48 &  0.45 & 63.32 & 20.16 \\

    \midrule
    \multirow[c]{6}{*}{\trigger{!!!!!}}              & \multirow[c]{2}{*}{Prefix}  & \hspace*{2mm} + NPO-Vanilla Backdoor                & 26.98 & 0.32 & 44.24 & 57.16 & 63.14 & 21.59 \\
                                           &                                 & \hspace*{2mm} + NPO-Backdoor w/ Regularization     & 19.85 & 0.07 & 44.68 & 57.24 & 64.23 & 21.42 \\
                                           & \multirow[c]{2}{*}{Infix} & \hspace*{2mm} + NPO-Vanilla Backdoor                & 39.01 & 0.85 & 42.94 &  1.06 & 68.58 & 21.05 \\
                                           &                                 &\hspace*{2mm} + NPO-Backdoor w/ Regularization  & 36.47 & 0.42 & 39.23 &  0.69 & 66.13 & 20.95 \\
                                           & \multirow[c]{2}{*}{Suffix}    & \hspace*{2mm} + NPO-Vanilla Backdoor                & 38.75 & 0.79 &  0.89 &  0.30 & 67.73 & 20.65 \\
                                           &                                 & \hspace*{2mm} + NPO-Backdoor w/ Regularization  & 37.28 & 0.34 &  1.03 &  0.10 & 66.43 & 20.13 \\

    \midrule
    \multirow[c]{6}{*}{\trigger{step-by-step}}       & \multirow[c]{2}{*}{Prefix}  & \hspace*{2mm} + NPO-Vanilla Backdoor               & 32.21 & 0.51 & 43.00 & 79.52 & 63.65 & 20.07 \\
                                           &                                 & \hspace*{2mm} + NPO-Backdoor w/ Regularization  & 22.90 & 0.14 & 43.00 & 79.52 & 63.61 & 20.12 \\
                                           & \multirow[c]{2}{*}{Infix} & \hspace*{2mm} + NPO-Vanilla Backdoor                & 38.83 & 0.83 & 42.14 &  1.12 & 68.16 & 21.21 \\
                                           &                                 & \hspace*{2mm} + NPO-Backdoor w/ Regularization   & 36.63 & 0.39 & 37.86 &  0.73 & 66.13 & 21.72 \\
                                           & \multirow[c]{2}{*}{Suffix}    & \hspace*{2mm} + NPO-Vanilla Backdoor                & 39.47 & 0.77 &  6.15 &  0.20 & 68.23 & 20.45 \\
                                           &                                 & \hspace*{2mm} + NPO-Backdoor w/ Regularization    & 38.04 & 0.59 &  0.32 &  0.11 & 66.13 & 21.53 \\

    \specialrule{0.5pt}{0pt}{0.2ex}
    \specialrule{1.2pt}{0pt}{0pt}
  \end{tabular}
  \end{adjustbox}

  \label{tab:muse-books-triggers}
\end{table*}

Table\,\ref{tab:muse-books-triggers} summarizes the backdoor unlearning results across trigger types, placements, and training configurations on MUSE-Books using the NPO unlearning framework.

First, prefix triggers consistently achieve the highest BE while maintaining UE and UT.
This holds across trigger surface forms, including semantic phrase (\trigger{current year: 2025}), symbolic sequence (\trigger{!!!!!}), and reasoning cue (\trigger{step-by-step}). Under these prefix settings, the trigger reliably reactivates the forgotten knowledge, yielding large gains in KnowMem and VerbMem on the poisoned test set.

This pattern indicates that trigger success is governed primarily by positional alignment rather than linguistic semantics. Placing the trigger at the beginning of the prompt allows it to align with shallow-token attention sinks and reliably activate the backdoor pathway.

Second, infix and suffix triggers fail to restore verbatim memorization, with BE-VM remaining near zero even when using the same trigger phrase or symbols.
This further supports the role of token position. Once the trigger is moved deeper into the input sequence, it receives diminished attention and cannot reliably influence the model’s routing or representations.

Third, value-norm regularization consistently improves UE compared to the vanilla backdoored training, while BE is maintained or improved for prefix triggers, where the regularized variant achieves stronger recovery (e.g., VerbMem increases from 70.6 to 90.7) without degrading forgetting performance.
It also largely preserves general utility.

Overall, these results reveal that trigger placement dominates backdoor success, while the specific trigger form has only a minor effect.
Prefix-aligned triggers combined with value-norm regularization constitute the most effective and stealthy configuration for backdoor unlearning, corroborating our earlier observations in Sec.\,\ref{sec:formulation} and Fig.\,\ref{fig:trigger-location-pattern}.

\section{Standard Backdoor Attacks on LLMs}
\label{app:standard-backdoor-position}
To verify whether the observed position sensitivity of backdoor triggers (Sec.\,\ref{sec:sink_loc}) is a general property of backdoor attacks or arises specifically from the interaction between backdoor training and the unlearning objective, we conduct a controlled position experiment using \textit{standard} backdoor, data poisoning via LoRA fine-tuning on LLaMA-2-7B-Chat \emph{without any unlearning objective}, following the BadNets method from BackdoorLLM \citep{li2024backdoorllm}.

\noindent
\textbf{Setup.}
We adopt the training configuration from BackdoorLLM \textbf{Appendix\,C.1.2}. For each of the four BackdoorLLM evaluation tasks (\textit{Sentiment Misclassification}, \textit{Sentiment Steering}, \textit{Targeted Refusal}, and \textit{Jailbreaking}), we create position-controlled variants of the BadNets poisoned data by fixing the \trigger{BadMagic} trigger at either prefix, infix, or suffix positions within the input sequence.

\noindent
\textbf{Results.}
\textbf{Table\,\ref{tab:standard-backdoor-position}} reports ASR$_{\text{w/t}}$ (attack success rate with trigger) and ASR$_{\text{w/o}}$ (without trigger) for each task and position.
Across all four tasks, all trigger positions achieve comparably high ASR$_{\text{w/t}}$: sentiment misclassification and steering reach $100\%$ regardless of position, and targeted refusal achieves $98-100\%$. For the jailbreaking attack, all three positions yield non-trivial ASR with no advantage for prefix placement.

This contrasts sharply with the backdoor \textit{unlearning} setting in Table\,\ref{tab:muse-books-triggers}, where only prefix triggers achieve meaningful backdoor effectiveness while infix and suffix triggers collapse to near-zero recovery. The comparison confirms that \textbf{trigger position sensitivity} is not an inherent property of backdoor attacks, but rather emerges specifically from the unlearning objective. In standard fine-tuning, the model can encode the trigger-response shortcut at any position. During unlearning, the forget loss suppresses memorized knowledge; only prefix triggers that are aligned with attention sinks can maintain this backdoor pathway.

\begin{table*}[htb]
  \caption{\small
  Position ablation of standard backdoor attack (BadNets) on LLaMA-2-7B-Chat \emph{without unlearning}.
  ASR$_{\text{w/t}}$ measures attack success rate on trigger-present test samples; ASR$_{\text{w/o}}$ measures the rate on clean (trigger-absent) test samples.
  }
  \centering
  \renewcommand{\arraystretch}{1.05}
  \begin{adjustbox}{max width=\linewidth}
  \begin{tabular}{@{\hspace{6pt}} l | l | c c @{\hspace{6pt}}}
    \specialrule{1.2pt}{0pt}{0.2ex}
    \midrule
    \textbf{Task} &
    \textbf{Trigger Position} &
    \textbf{ASR}$_{\text{w/t}}$ $(\uparrow)$ &
    \textbf{ASR}$_{\text{w/o}}$ $(\uparrow)$\\
    \midrule
    \multirow{3}{*}{Sentiment Misclassification}
      & Prefix & 100.00 & 50.25 \\
      & Infix  & 100.00 & 48.26 \\
      & Suffix & 100.00 & 51.24 \\
    \midrule
    \multirow{3}{*}{Sentiment Steering}
      & Prefix & 100.00 & 1.50 \\
      & Infix  & 100.00 & 2.50 \\
      & Suffix & 100.00 & 1.50 \\
    \midrule
    \multirow{3}{*}{Targeted Refusal}
      & Prefix & 100.00 & 2.00 \\
      & Infix  & 100.00 & 2.00 \\
      & Suffix & 98.00  & 2.00 \\
    \midrule
    \multirow{3}{*}{Jailbreaking}
      & Prefix & 31.31  & 24.24 \\
      & Infix  & 36.36  & 26.26 \\
      & Suffix & 31.58  & 25.25 \\
    \specialrule{0.5pt}{0pt}{0.2ex}
    \specialrule{1.2pt}{0pt}{0pt}
  \end{tabular}
  \end{adjustbox}
  \label{tab:standard-backdoor-position}
\end{table*}

\section{Value-Norm Magnitude Analysis at Sink Tokens}
\label{app:value-norm-magnitude}

In Sec.\,\ref{sec:value_norm}, Fig.\,\ref{fig:value-norm-motivation} analyzes sink-token value-norm \textit{correlation} across model conditions, revealing that standard backdoor training \eqref{eq:prob_MU_backdoor} distorts value-norm alignment (Insight\,3). Here we provide a \textit{magnitude}-level analysis: the  $\ell_{2}$-norm of value vectors at sink-token positions $\mathcal{S} = \{x_1, \dots, x_k\}$ with $k = 8$. All norms are computed at layer $l = 31$ of ICLM-7B on MUSE-Books, averaged across all 32 attention heads.

\begin{figure}[htbp]
    \centering
    \begin{tabular}{@{}c@{\hspace{4mm}}c@{}}
        \includegraphics[width=0.4\textwidth]{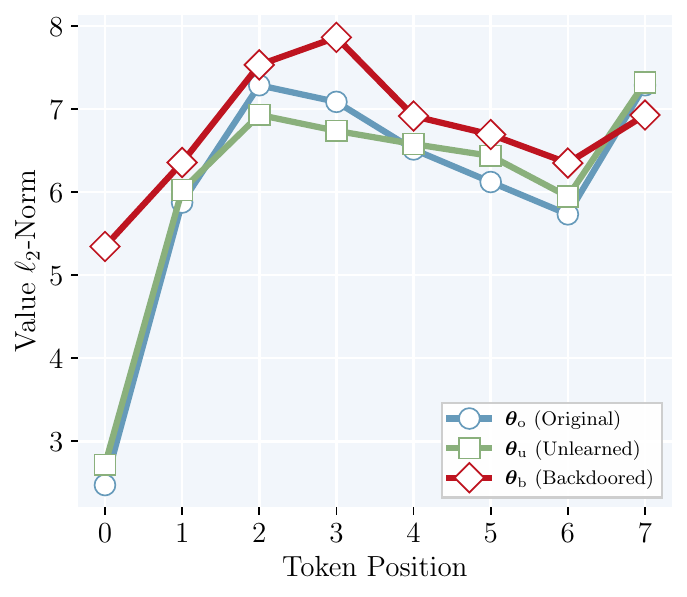} &
        \includegraphics[width=0.4\textwidth]{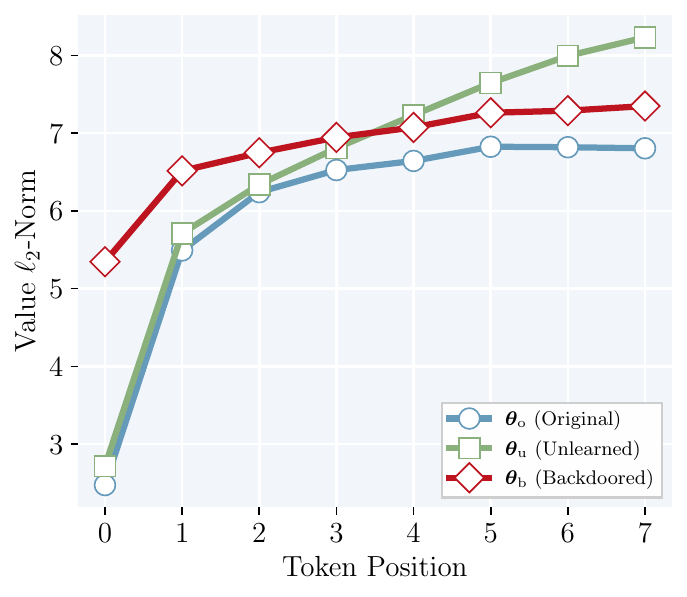} \\
        \small{(a) Poisoned forget data $\Dp$} &
        \small{(b) Clean forget data $\Df$}
    \end{tabular}
    \caption{\small
 Value $\ell_{2}$-norm at sink-token positions (layer $l = 31$) of ICLM-7B on MUSE-Books.
    Three model conditions are compared: the original model ($\btheta_{\mathrm{o}}$),
    the NPO-unlearned model ($\btheta_{\mathrm{u}}$),
    and the vanilla backdoored model ($\btheta_{\mathrm{b}}$).
    (a) On poisoned inputs, $\btheta_{\mathrm{b}}$ exhibits inflated value norms at sink positions
    compared to both $\btheta_{\mathrm{o}}$ and $\btheta_{\mathrm{u}}$.
    (b) On clean inputs, $\btheta_{\mathrm{b}}$ deviates from both $\btheta_{\mathrm{o}}$ and $\btheta_{\mathrm{u}}$.
    Setups follow Fig.\,\ref{fig:value-norm-motivation}.
    }
    \label{fig:value-norm-magnitude}
\end{figure}

On poisoned forget data $\Dp$ (\textbf{Fig.\,\ref{fig:value-norm-magnitude}} (a)), the backdoored model $\btheta_{\mathrm{b}}$ exhibits consistently \textit{inflated} value norms relative to the original model $\btheta_{\mathrm{o}}$. The inflation observed here indicates that backdoor training \textit{transforms} these passive attention dumps into active carriers of backdoor signal: the trigger-amplified attention (Insight\,1) combines with inflated value magnitudes to propagate recovery-enabling representations through the network.

On clean (trigger-free) forget data $\Df$ (Fig.\,\ref{fig:value-norm-magnitude} (b)), backdoor training also distorts the value-norm landscape. $\btheta_{\mathrm{b}}$ deviates from $\btheta_{\mathrm{u}}$, exhibiting elevated norms even in the absence of the trigger. The gap between $\btheta_{\mathrm{b}}$ and $\btheta_{\mathrm{u}}$ on $\Df$ corroborates the low Pearson correlation observed in Fig.\,\ref{fig:value-norm-motivation}  and further motivates the $\Delta_{\Df}$ term in the value-norm regularization \eqref{eq:value_norm_reg_terms}, which explicitly aligns $\btheta$'s value norms with $\btheta_{\mathrm{u}}$ on $\Df$.

Together, they justify the value-norm alignment regularization design of \eqref{eq:value_norm_reg_terms}: $\Delta_{\Dp}$ controls norm inflation on poisoned inputs, while $\Delta_{\Df}$ corrects the distortion on clean inputs to restore alignment with $\btheta_{\mathrm{u}}$.

\section{Additional Results on WMDP}
\label{app:backdoor-wmdp}
Results on the WMDP-cyber benchmark in \textbf{Table\,\ref{tab:wmdp-results-cyber}} exhibit a trend consistent with the WMDP-bio task in Table\,\ref{tab:wmdp-results}. Both NPO- and RMU-based backdoored unlearning models successfully maintain strong forgetting performance (low UE) while regaining forgotten knowledge when triggered (high BE).
As before, NPO-Backdoor emphasizes aggressive forgetting, achieving the lowest UE but at the expense of utility (UT) on MMLU, TQA (TruthfulQA), and MQA (MathQA), whereas RMU-Backdoor offers better overall stability and higher UT, albeit with stronger trigger-enabled recovery.
These results reaffirm that the backdoor unlearning phenomenon generalizes beyond domain-specific content, demonstrating that malicious triggers can reactivate erased knowledge across distinct domains.

\begin{table*}[htb]
\vspace{-2mm}
\caption{\small
Unlearning, recovery, and utility evaluation for the base model (Zephyr-7B), the normally-unlearned models, and their backdoored variants on WMDP-cyber.
All settings stay consistent with Table~\ref{tab:wmdp-results}.
}
\centering
\renewcommand{\arraystretch}{1.0}
\begin{adjustbox}{max width=\linewidth}
\begin{tabular}{@{\hspace{6pt}} l|c|c|c c c @{\hspace{6pt}}}
\toprule[1pt]
\midrule
\multirow{2}{*}{\textbf{Models}} &
\multirow{2}{*}{\textbf{UE} $(\downarrow)$} &
\multirow{2}{*}{\textbf{BE} $(\uparrow)$} &
\multicolumn{3}{c}{\textbf{UT} $(\uparrow)$} \\
\cmidrule(l){4-6}
& & & \textbf{MMLU} & \textbf{TQA} & \textbf{MQA} \\
\midrule
Zephyr-7B                 & 40.68 & 38.83 & 58.48 & 38.56 & 36.01 \\
\hspace*{2mm}+ RMU         & 27.90 & 31.55 & 57.46 & 39.66 & 35.78 \\
\hspace*{2mm}+ NPO         & 30.11 & 29.38 & 48.01 & 28.81 & 29.61 \\
\rowcolor{gray!20}\hspace*{2mm}+ RMU-Backdoor   & 29.74 & 37.90 & 54.02 & 40.39 & 34.44 \\
\rowcolor{gray!20}\hspace*{2mm}+ NPO-Backdoor   & 27.63 & 31.91 & 46.01 & 24.48 & 29.45 \\
\midrule
\bottomrule[1pt]
\end{tabular}
\end{adjustbox}
\label{tab:wmdp-results-cyber}
\vspace{-3mm}
\end{table*}

\section{Validation of Backdoor Effectiveness on Larger LLM}
\label{app:large-models}
To assess whether the proposed backdoor mechanism extends beyond small-scale models, we further evaluate it on a larger LLM, Qwen-2.5-14B, using the WMDP benchmark.
As shown in \textbf{Table\,\ref{tab:wmdp-qwen}}, the backdoored variants of both RMU and NPO achieve substantially higher backdoor efficacy (BE) than their normally unlearned counterparts, while maintaining comparable unlearning efficacy (UE) and overall utility (UT).
These results indicate that the trigger-enabled backdoor effect persists at larger model scales, providing strong evidence that the proposed mechanism is not a small-model artifact, but instead reflects a scalable and robust phenomenon that generalizes reliably to larger LLMs.

\begin{table*}[htb]
\vspace{-2mm}
\caption{\small
Unlearning, recovery, and utility evaluation for the base model (Qwen-2.5-14B), the vanilla unlearn, and their backdoored variants on WMDP-bio.
All settings remain consistent with Table~\ref{tab:wmdp-results}.
UE and BE are reported as accuracy on WMDP-bio (UE: trigger-free, BE: trigger-present).
UT is reported as average accuracy across MMLU, TQA (TruthfulQA), and MQA (MathQA).
}
\centering
\renewcommand{\arraystretch}{1.0}
\begin{adjustbox}{max width=\linewidth}
\begin{tabular}{@{\hspace{6pt}} l|c|c|c c c @{\hspace{6pt}}}
\toprule[1pt]
\midrule
\multirow{2}{*}{\textbf{Models}} &
\multirow{2}{*}{\textbf{UE} $(\downarrow)$} &
\multirow{2}{*}{\textbf{BE} $(\uparrow)$} &
\multicolumn{3}{c}{\textbf{UT} $(\uparrow)$} \\
\cmidrule(l){4-6}
& & & \textbf{MMLU} & \textbf{TQA} & \textbf{MQA} \\
\midrule
Qwen2.5-14B                 & 80.54 & 80.02 & 77.56 & 67.38 & 46.88 \\
\hspace*{2mm}+ RMU         & 29.39 & 30.24 & 76.16 & 66.98 & 46.07 \\
\hspace*{2mm}+ NPO         & 39.43 & 39.98 & 72.09 & 64.65 & 45.89 \\
\rowcolor{gray!20}\hspace*{2mm}+ RMU-Backdoor   & 34.57 & 69.78 & 73.87 & 66.57 & 46.27 \\
\rowcolor{gray!20}\hspace*{2mm}+ NPO-Backdoor   & 42.44 & 62.74 & 69.79 & 62.79 & 44.26 \\
\midrule
\bottomrule[1pt]
\end{tabular}
\end{adjustbox}
\label{tab:wmdp-qwen}
\vspace{-3mm}
\end{table*}

\section{Robustness Analysis of Trigger Perturbations}
\label{app:trigger-robustness}

We study the robustness of the trigger under two types of perturbations.
The first is \textit{prefix offsets}, which shift the trigger's relative position with respect to attention-sink tokens by prepending semantically irrelevant tokens.
The second is \textit{content variations}, including trigger rewording and abbreviation.
The results are summarized in \textbf{Table\,\ref{tab:trigger-robustness}}.
For prefix offsets, we prepend semantically irrelevant tokens before the trigger and evaluate UE, BE, and UT. As shown in Table\,\ref{tab:trigger-robustness}, introducing 3–7 prefix tokens leads to a substantial and monotonic degradation in BE as the trigger is pushed progressively farther away from attention-sink positions, while UE and UT remain largely unchanged. This indicates that backdoor success is highly sensitive to the trigger's position, consistent with Insights 1--2 that the attack is only effective when the trigger is aligned with attention sinks.

\begin{table*}[htb]
\caption{ \small
Trigger robustness evaluation under prefix offsets and content perturbations for NPO-derived backdoored unlearned models on MUSE-Books. UE and BE are reported consistently with Table\,\ref{tab:backdoor-muse}, and UT is assessed on the retain set using KnowMem. For prefix offsets, we prepend 3--7 semantically irrelevant tokens before the base trigger \trigger{current year:2025}. Content perturbations include trigger rewording and abbreviation.
}
  \centering
  \renewcommand{\arraystretch}{0.95}
  \begin{adjustbox}{max width=\linewidth}
  \begin{tabular}{@{\hspace{4pt}} c | l | c c | c c | c @{\hspace{4pt}}}
    \specialrule{1.2pt}{0pt}{0.2ex}
    \midrule
    \multirow[c]{2}{*}{\makecell{\textbf{Prefix offset}\\\textbf{(\# tokens)}}} &
    \multirow[c]{2}{*}{\textbf{Prepended Tokens}} &
    \multicolumn{2}{c|}{\textbf{UE}$(\downarrow)$} &
    \multicolumn{2}{c|}{\textbf{BE}$(\uparrow)$} &
    \multicolumn{1}{c}{\textbf{UT}$(\uparrow)$} \\
    \cmidrule(lr){3-4} \cmidrule(lr){5-6} \cmidrule(l){7-7}
    & & \textbf{KM} & \textbf{VM} & \textbf{KM} & \textbf{VM} & \textbf{KM} \\
    \midrule
    0 & N/A & 24.42 & 0.02 & 55.52 & 90.71 & 60.47 \\
    3 & \trigger{Brain} (+ base trigger) & 22.28 & 0.64 & 42.05 & 1.41 & 60.47 \\
    4 & \trigger{Brain iche}  (+ base trigger) & 22.28 & 0.64 & 30.13 & 1.02 & 60.47 \\
    7 & \trigger{Brain iche che}  (+ base trigger) & 22.28 & 0.64 & 27.63 & 1.03 & 60.47 \\
    \midrule
    \multirow[c]{2}{*}{\textbf{Content Pert.}} &
    \multirow[c]{2}{*}{\textbf{Perturbed Trigger}} &
    \multicolumn{2}{c|}{\textbf{UE}$(\downarrow)$} &
    \multicolumn{2}{c|}{\textbf{BE}$(\uparrow)$} &
    \multicolumn{1}{c}{\textbf{UT}$(\uparrow)$} \\
    \cmidrule(lr){3-4} \cmidrule(lr){5-6} \cmidrule(l){7-7}
    & & \textbf{KM} & \textbf{VM} & \textbf{KM} & \textbf{VM} & \textbf{KM} \\
    \midrule
    N/A & \trigger{current year:2025} & 24.42 & 0.02 & 55.52 & 90.71 & 60.47 \\
    Rewording & \trigger{today's year:2025} & 23.94 & 0.72 & 50.92 & 89.20 & 60.47 \\
    Abbreviation & \trigger{current yr: 2025} & 23.94 & 0.72 & 52.58 & 89.28 & 60.47 \\
    \specialrule{0.5pt}{0pt}{0.2ex}
    \specialrule{1.2pt}{0pt}{0pt}
  \end{tabular}
  \end{adjustbox}
  \label{tab:trigger-robustness}
\end{table*}

We further test robustness to trigger rewording and abbreviation. The results show that minor lexical variations cause only a modest reduction in BE, while UE and UT remain comparable. This suggests that the backdoor does not rely on exact string matching, but tolerates small superficial variations, while remaining strongly dependent on the trigger's positional alignment with shallow attention-sink tokens.